# Applications of Probabilistic Programming

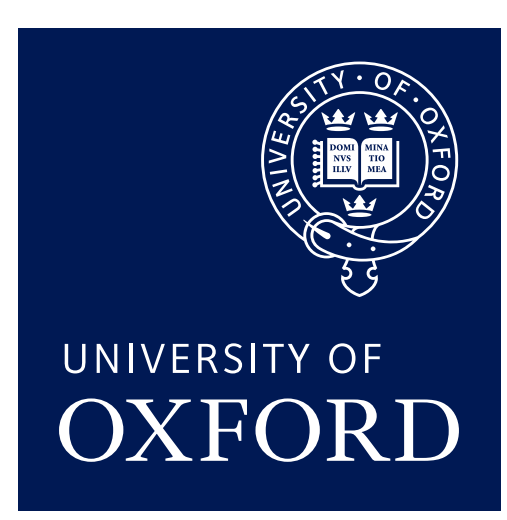

Yura Perov

Wolfson College

Department of Engineering Science

University of Oxford

A thesis submitted for the degree of

*Master of Science by Research*

Trinity 2015

# Abstract

This thesis describes work on two applications of probabilistic programming: the learning of probabilistic program code given specifications, in particular program code of one-dimensional samplers; and the facilitation of sequential Monte Carlo inference with help of data-driven proposals. The latter is presented with experimental results on a linear Gaussian model and a non-parametric dependent Dirichlet process mixture of objects model for object recognition and tracking.

We begin this work by providing a brief introduction to probabilistic programming.

In the second Chapter we present an approach to automatic discovery of samplers in the form of probabilistic programs. Specifically, we learn the procedure code of samplers for one-dimensional distributions. We formulate a Bayesian approach to this problem by specifying a grammar-based prior over probabilistic program code. We use an approximate Bayesian computation method to learn the programs, whose executions generate samples that statistically match observed data or analytical characteristics of distributions of interest. In our experiments we leverage different probabilistic programming systems, including Anglican and Probabilistic C, to perform Markov chain Monte Carlo sampling over the space of programs. Experimental results have demonstrated that, using the proposed methodology, we can learn approximate and even some exact samplers. Finally, we show that our results are competitive with regard to genetic programming methods.

In Chapter 3, we describe a way to facilitate sequential Monte Carlo inference in probabilistic programming using data-driven proposals. In particular, we develop a distance-based proposal for the non-parametric dependent Dirichlet process mixture of objects model. We implement this approach in the probabilistic programming system Anglican, and show that for that model data-driven proposals provide significant performance improvements. We also explore the possibility of using neural networks to improve data-driven proposals.

# Heartfelt gratitude

There are so many beings and organisations who made this work possible.

First of all, I overwhelmingly thank my supervisor, Prof Frank Wood. His support as a supervisor is extraordinary. He helped me grow my initial idea seeds into fairly mature research projects, and helped me improve my research skills. He also has significantly supported and contributed to my overall personal development. I thank him extremely much.

I am grateful to my Oxford colleagues and mentors: Tuan Anh Le, Brooks Paige, Neil Dhir, Tom Rainforth, Dr David Tolpin, Dr Jan Willem van der Meent, Tom Jin, Niclas Palmius, David Janz, as well as to senior colleagues, including Prof Hongseok Yang, Prof Michael Osborne, Prof Steven Roberts, Prof Nando de Freitas, Prof Yee Whye Teh, Prof François Caron, and many others. I am appreciative of feedback from and discussions with members of the probabilistic programming reading group, the machine learning lunches and the brainstorming artificial intelligence forum sessions.

I thank Willie Neiswanger, whose help with the DDPMO model was very useful for our work with Tuan Anh Le and who also has provided us with some handy code that was important for the work.

I am appreciative of the support of the University, of my college, Wolfson, and also that of Somerville college, St Antony's college and Exeter college.

This thesis is the natural continuation of my work on probabilistic programming at the Massachusetts Institute of Technology. I thank my colleagues at the Massachusetts Institute of Technology, where the engagement with probabilistic programming and machine learning magic in general happened to me first time. I thank Dr Vikash Mansinghka and Prof Joshua Tenenbaum in particular, who made my visit to MIT, first of all, possible (thank you both very much for believing in me!), secondly, made it very fruitful in terms of improving my skills, and, last but not least, they made me feel it very cordially welcomed. I am extremely grateful to Vikash for all enormous amount of time and effort he committed to supervising and working with me. May I also thank my colleagues at MIT, in particular Tejas Kulkarni, Ardavan Saeedi, Daniel Selsam, Andreas Stuhlmüller, Jonathan Huggins, Jonathan Malmaud, Dan Lovell, Jay Baxter, Zack Drach, Vlad Firoiu, Max Siegel, Max Kleiman-Weiner, Dr Cameron Freer, and many others, very much.

I would also like to thank my colleagues and mentors at the Siberian Federal University, and in Russia in general, who educated me in Mathematics, Computer Science and Economics, and who also made my studies, as a visiting student, at MIT and Oxford, possible. In particular, I thank Prof Alexander Gorban (who is currently professor at the University of Leicester), Prof Alexander Kitmanov, Dr Tatiana Krupkina, Prof Eugene Semenkin, Vladimir Kontorin, Tatiana Mihailova, and literally hundreds of others. I am also appreciative of advice and support I received from the members of Russian Association for Artificial Intelligence, including Prof Vadim Stefanuk, Prof Gennady Osipov, Prof Oleg Kuznezov, Prof Igor Fominih, Prof Vladimir Khoroshevskij, Prof Vadim Vagin, and others.

Everything I do, in terms of good and productive things, is made possible because of my beloved relatives and friends. I am extremely grateful to my parents, Olga and Nikolay, to my sister Olga and to her husband Igor, to my grandmas Galia and Toma, to my aunt Natalja, and, of course, to Maria. I thank Ardavan, Tejas, Daniil, Michelle, Neil and Roman for their huge friendly support.

This work, in particular that on learning probabilistic programs, and my skills have also significantly benefited from meetings and discussions with other research groups and researchers, including Microsoft Research at Cambridge (in particular Dr Andy Gordon, Dr Ali Eslami, Dr John Winn, Dr Tom Minka; and others), Microsoft Research at Bangalore (Dr Aditya Nori), Stanford (Prof Percy Liang, Prof Noah Goodman, and their research groups), Cambridge (Prof Zoubin Ghahramani, and his group), Berkley (Prof Stuart Russell, Prof Rastislav Bodik, Prof Tom Griffiths, and their groups), Harvard (Prof Ryan Adams and his group), École Polytechnique Fédérale de Lausanne (Prof Patrick Thiran, Prof Matthias Grossglauser, Prof Auke Ijspeert and his group), the Dagstuhl seminar on "Approaches and Applications of Inductive Programming" (December 2013), and the meetings at the Heidelberg Laureate Forum (August 2015).

May I also thank the Supreme Being, and also all beings on this planet. I believe that I might be thankful to almost every person on this planet who have been helping me with my research work and my studies in this or that way though the huge interconnection of the world economy today.

These few pages are definitely not enough to list all people, whose help, mentoring, support and discussions have led me to be at Oxford and allowed me to work on these projects. Thank you, all of you, very much!

I am really grateful to all sponsors who made this work possible, including the Russian President's fellowship, Xerox, Google, DARPA, Somerville College, my relatives, incomes and savings from my previous business projects (thus I thank clients, partners, investors and other contributors of those projects), and help from my friends.

# Acknowledgement

The initial description of the approach and initial experimental results on learning probabilistic programs were described in an arXiv submission (Perov and Wood, 2014), and in my Bachelor's thesis (Perov, 2014).

This is also to acknowledge that the part of the work on data-driven proposals was made in collaboration with Tuan Anh Le under the supervision of Prof Frank Wood. In particular, we brainstormed and developed the ideas together, wrote probabilistic program code for linear Gaussian model and DDPMO model experiments, and performed and analysed sets of experiments together.

The proposal on the experiment for scientific rediscovery of classical genetics laws was prepared with advice of Prof Joshua Tenenbaum.

For initial experiments, a framework for Stochastic Simulation in Java (L'Ecuyer et al., 2002) was usefully employed.

# Chapter 1

# Introduction to probabilistic programming

Probabilistic programming (Goodman, 2013; Gordon et al., 2014; De Raedt and Kimmig, 2013; Ranca, 2014) is a constructivist way to describe probabilistic models and conduct statistical inference in such models given data. A probabilistic model in the form of probabilistic program code describes the data-generating process from unknown latent variables' values to observed data. The latent variables values are subject to assumptions that are given in the form of probability distributions. The types of models that may be written as probabilistic programs include not only basic Bayesian networks and graphical models, but also ones more expressive and flexible, such as non-parametric models and graphical models with dynamic structure.

## 1.1 An example of a probabilistic program, and its execution trace

A basic probabilistic model in the form of a probabilistic program is shown below:

```
(query
  (let
    [unknown-mean-t1 (sample (normal 2 1))
     unknown-mean-t2 (sample (normal unknown-mean-t1 1))
     noise 0.1]
    (observe (normal unknown-mean-t1 noise) 3)
    (observe (normal unknown-mean-t2 noise) 3.1)
    (predict unknown-mean-t1)
    (predict unknown-mean-t2)))
```

Figure 1.1: A probabilistic model with Gaussian emissions with unknown means and a known standard deviation. The model is given in probabilistic programming language Anglican (Tolpin et al., 2015b).

The accompanying statistical model can be expressed as follows:

$$
\begin{aligned}
x_1 &\sim \text{Normal}\,(2, 1)\,, \\
x_2 &\sim \text{Normal}\,(x_1, 1)\,, \\
y_1 \mid x_1 &\sim \text{Normal}\left(x_1, 0.1^2\right), \\
y_2 \mid x_2 &\sim \text{Normal}\left(x_2, 0.1^2\right).
\end{aligned}
\tag{1.1}
$$

Equation 1.1 contains two latent variables, $x_1$ and $x_2$, and two observed data points, $y_1$ and $y_2$. Each run of a probabilistic program yields a single execution trace. An execution trace is a map from random choices to their specific values. An execution trace fully defines the execution of the probabilistic program. Given the execution trace, a probabilistic program becomes deterministic. An example of the execution trace for the program in Figure 1.1 is $(x_1 = 3.0, x_2 = 2.5, y_1 = 2.0, y_2 = 2.1)$. For that particular program, each list of four random variables constitutes a valid execution trace: $(x_1 \in \mathbb{R}, x_2 \in \mathbb{R}, y_1 \in \mathbb{R}, y_2 \in \mathbb{R})$.

The probability of an execution trace can be defined, in a similar but more restrictive way to (Wood et al., 2014), as $p(\mathbf{y}, \mathbf{x}) \equiv \prod_{n=1}^{N} p(y_n|\zeta_{t_n}, \mathbf{x}_n)p(\mathbf{x}_n|\mathbf{x}_{n-1})$, where $y_n$ is the $n$-th output data point (i.e. an observation), $p(y_n|\zeta_{t_n}, \mathbf{x}_n)$ is its normalised likelihood, $\zeta_{t_n}(\mathbf{x}_n)$ is its argument, $t_n(\mathbf{x}_n)$ is a random procedure type (e.g. Normal), $\mathbf{x}_n$ is the ordered set of all random choices that have to be computed before the likelihood of $y_n$ can be evaluated, $p(\mathbf{x}_n|\mathbf{x}_{n-1})$ is its normalised prior probability, and $\mathbf{x}$ and $\mathbf{y}$ are the sets of all latent and observing random procedure applications correspondingly.

## 1.2 Another example of a probabilistic program, and the related execution trace

Another simple example of a probabilistic program is a procedure that samples from the geometric distribution:

```
(query
  (let
    [geometric
      (fn geometric [p]
         (if (sample (flip p))
           1
           (+ (geometric p) 1)))]
    (predict (geometric 0.3))))
```

Figure 1.2: A probabilistic program that samples from the prior of the Geometric distribution with parameter $0 < p < 1$. This probabilistic program and its trace has a varying number of random choices.

For simplicity, this second example of a probabilistic program does not have any observations, and thus it is unconditioned. On other words, its prior is the same as the posterior.

Examples of valid execution traces of this program include:

1. `true;`

2. `false, true;`

3. `false, false, false, false, true.`

A valid execution trace for that program is any, possibly empty as well, sequence of false draws, terminated by a true draw. The program is a good example, because it demonstrates that while the number of drawn random choices in any execution trace is expected to be always finite for programs that terminate with probability $1$, that number is not necessarily bounded by any constant.

## 1.3 Existing probabilistic programming platforms and statistical inference in them

Many probabilistic programming languages have been designed. Usually, for each probabilistic programming language, a new related engine is developed to conduct some type of statistical inference. Statistical inference methods are typically chosen and implemented in such a way as to allow automatic inference for any probabilistic programs that may be expressed in a probabilistic programming language. The expressiveness of the language thus varies, and depends on utilised inference methods.

Particular languages and implementations include functional probabilistic programming languages such as Church (Goodman et al., 2008), Anglican (Wood et al., 2014) and Venture (Mansinghka et al., 2014); logic probabilistic programming languages (De Raedt and Kimmig, 2013) such as ProbLog (Kimmig et al., 2011); and domain-specific PPLs, such as (Kiselyov and Shan, 2009). Other languages and implementations also endorse declarative definitions of probabilistic models and include IBAL (Pfeffer, 2001),

Stan (Stan Development Team, 2014), BLOG (Milch et al., 2007), BUGS (Lunn et al., 2009), FACTORIE (McCallum et al., 2009), Markov Logic networks (Richardson and Domingos, 2006), and Infer.NET (Minka et al., 2012). More detailed overviews are given in (Roy, 2016; De Raedt and Kimmig, 2013; Gordon et al., 2014; Mansinghka et al., 2014).

These implementations employ different statistical inference methods, which include Markov chain Monte Carlo (Milch et al., 2007; Goodman et al., 2008; Mansinghka et al., 2014; Lunn et al., 2009; Milch et al., 2007), sequential Monte Carlo (Wood et al., 2014), Hamiltonian Monte Carlo (Stan Development Team, 2014), variational inference (Mansinghka et al., 2014), belief propagation (Hershey et al., 2012), expectation propagation (Minka et al., 2012), and variational message passing (Minka et al., 2012).

The choice of the employed inference method is related to the trade-off between expressiveness and inference performance. For example, Infer.NET is one of the most high-performance probabilistic programming engines and may process huge datasets. This high performance is, however, achieved by internal compilation of an Infer.NET probabilistic program into a finite graphical model and an application of expectation propagation inference method (Minka, 2001), which seriously restricts the range of models that may be written in it. In particular, it is not possible to perform inference for non-parametric Bayesian models in Infer.NET. On the other hand, while languages like Church, Anglican and Venture are some of the most flexible and expressive (Ranca, 2014), their statistical inference performance is slower by at least the factor of 10x in comparison to languages like Infer.NET and at least by the factor of 50x in comparison to hand-written samplers. Here, by hand-written samplers we mean the implementations of inference with manually derived updated, written for specific models in fast languages such as C or C++.

## 1.4 Ways to improve general-purpose statistical inference in probabilistic programming platforms

Such slow performance means that one of the current critical drawbacks of probabilistic programming is the lack of efficient general-purpose inference algorithms. This problem prevents probabilistic programming from being utilised on a large scale by users in the machine learning field, including researchers, scientists, data analysts and graduate students. Several possible approaches are being explored to address this issue. One is to employ new general-purpose inference methods. For example, two new probabilistic programming languages have recently been introduced, employing sequential Monte Carlo methods (Smith et al., 2013): Anglican (Wood et al., 2014) and Biips (Todeschini et al., 2014). In 2014 a general-purpose implementation of the particle Gibbs with ancestor sampling method (Lindsten et al., 2014) was introduced in (van de Meent et al., 2015) as an alternative engine for Anglican. Variational inference has been employed in probabilistic programming since 2013 in Stochastic MATLAB (Wingate and Weber, 2013), Venture (Mansinghka et al., 2014) and Stan (Kucukelbir et al., 2014). Finally, slice sampling for probabilistic programming has been proposed in (Ranca and Ghahramani, 2015).

In addition to statistical inference methods, optimisation methods have been applied for probabilistic programming. In particular, an approximation search algorithm for maximum a posteriori probability estimation has been presented in (Tolpin and Wood, 2015).

Another approach to facilitate inference for probabilistic programming is to increase the performance of already employed inference methods. Foremost, this is achieved by technical enhancements in implementations of probabilistic programming engines: by the choice of a faster implementation language (e.g. the implementation of Venture in C++ (Mansinghka et al., 2014) works faster than its very early prototype implementation in Clojure (Perov and Mansinghka, 2012)), by an intermediate compilation instead of a continuous interpretation (e.g. the recent Anglican implementation (Tolpin et al., 2015b), with a program compilation into a Clojure function, works faster than the previous Anglican interpreter (Wood et al., 2014)), and by an utilisation of just-in-time compilation (examples include engines, described in (Perov and Mansinghka, 2012; Tolpin et al., 2015b), which are implemented in Clojure). Another related work worth mentioning is "Probabilistic C" (Paige and Wood, 2014), where authors present a C library that allows sequential Monte Carlo and Particle Gibbs inference in any C and C++ program with just two added C functions, `OBSERVE` and `PREDICT`, to condition executions and get particle smoothing predictions correspondingly. This not only provides a very fast probabilistic programming engine implementation, but also serves as a compilation target and allows the transformation of almost any existing deterministic programming language into one that is probabilistic. Mostly, all these mentioned methods give constant performance improvements in time and memory. Furthermore, ways have been proposed to enhance the design of existing inference algorithms, e.g. by exploring conditional dependencies and making incremental updates only on a part of the execution trace, as in Venture (Perov and Mansinghka, 2012; Mansinghka et al., 2014) and Shred (Yang et al., 2014), where the latter is a tracing interpreter for Church language. For many models, these methods give asymptotic performance improvements in execution time. For example, while in old implementations of Church (Goodman et al., 2008) $N$ sweeps[1] of Metropolis-Hastings inference in a hidden Markov model with $T$

[1] A sweep, in the context of doing Metropolis-Hastings inference on the probabilistic program with $T$ random choices, consists of $T$ local MH proposals on those random choices. One local MH proposal on a random choice is when we propose a new value just for one random choice

data points had time complexity of $O(T^2N)$, in Venture[2] this complexity is $O(TN)$.

So far, we discussed two approaches to improve probabilistic programming inference, exploitation of new general-purpose inference algorithms and their variations, and the increase of performance of already employed algorithms, without changing their statistical convergence properties. Another approach is to improve the statistical properties of inference algorithms such that they converge faster. An example of this approach is the improvement of message passing in expectation propagation (EP) by learning message passing operators. Methods to learn EP operators for inference in Infer.NET have included neural networks (Heess et al., 2013), just-in-time random forests (Eslami et al., 2014) and just-in-time kernel-based regression (Jitkrittum et al., 2015). In addition, to improve general-purpose message passing in Infer.NET for graphical models with many layers, which are often used in computer vision, a consensus message passing method has been proposed in (Jampani et al., 2015a). Another example of enhancing general-purpose inference is a recent paper (Tolpin et al., 2015a) introducing *adaptive* Metropolis-Hastings inference for probabilistic programming. That work proposes to learn an MCMC scheduler: they learn non-uniform probabilities of proposing random variables, on which the engine makes proposals. Their results show that their MCMC scheduler provides a consistent improvement in convergence. The way on using data-driven proposals and discriminative models described in Chapter 3 of this work is related to this approach.

Two following chapters illustrate examples of what problems might be addressed

(given the fixed values of all other random choices in the execution trace), and accept or reject its new value.

[2] Venture has such efficient asymptotics not only for MH, but for versions of Gibbs inference algorithm as well. It is also capable of handling efficient inference in probabilistic programs with varying number of random choices.

with the help of the probabilistic programming framework. In particular, Chapter 2 describes an approach to learning probabilistic programs automatically, using existing highly expressive probabilistic programming platforms that support higher-order functions. The ultimate goal of work described in Chapter 2 is to automatically induce and employ generative models of the world for general artificial intelligence.

Chapter 3 demonstrates how to facilitate statistical inference in probabilistic programming by using discriminative models in order to improve Monte Carlo proposals. The approach is illustrated by experiments on the existing Bayesian generative nonparametric model, "the Dependent Dirichlet Process Mixture of Objects" (Neiswanger et al., 2014). While this model has already existed in the field of machine learning, this is the first time it has been implemented in a probabilistic programming framework.

# Chapter 2

# Learning probabilistic programs

The aim of many machine learning algorithms is to process existing data and provide predictions. To do so, model parameters and/or model structure need to be learnt.

In this Chapter we present an approach to automatic discovery of generative models in the framework of probabilistic programming. Here, probabilistic programing is a suitable approach, since a probabilistic program is, in essence, a procedural representation of a generative model. The ultimate goal for the future will be to automatically induce generative models for the general artificial intelligence.

Our intermediate task is far more modest. In this thesis, we aim to induce program code that, when executed repeatedly, returns values the distribution of which matches that of observed data. As a starting point, we consider the induction of programs that sample from parametrised one-dimensional distributions. In other words, the problem is to automatically learn simple versions of generative models (samplers), that statistically match observed data. Such samplers are to be learnt in the form of potentially interpretable probabilistic program code.

Probabilistic programming is relevant to this problem because programs in Turing-complete languages can represent a wide range of generative probabilistic models, and samples from these models can be generated efficiently by simply executing the program code. Representing generative models as code has the additional advantage that learned programs may potentially be analysed by humans. Finally, in higher-order languages like Anglican, where procedures may act on other procedures, it is possible to write a generative model for program code that is itself a probabilistic program. This enables us to perform inference by specifying an adaptor-based grammar prior over program code and to use general-purpose Markov chain Monte Carlo algorithms implemented by the inference engine of Anglican (Tolpin et al., 2015b) to sample over the space of programs.

To assess whether the distribution of samples generated by a program candidate matches the given distribution of interest, we use approximate Bayesian computation methods (Marin et al., 2012). We specify an approximate likelihood in terms of the similarity between a summary statistic of the generated samples and that of the observed distribution of interest. While this approach is inherently approximate, it still can be used to find exact sampler code. This argument is supported by the fact that we were able to successfully learn an exact sampler for the Bernoulli distribution family (Perov and Wood, 2014; Perov, 2014), given only an adaptor grammar-based prior learnt from a corpus of sampler code that did not include Bernoulli sampler code. We also found approximate samplers for other common one-dimensional distributions and for real-world data. Finally, our approach holds its own in comparison to state-of-the-art genetic programming methods (Koza, 1992; Poli et al., 2008).

The probabilistic programming language and system we use, Anglican, is Turing-complete and higher-order, and this allows us to specify the grammar prior as a higher-level probabilistic program that samples probabilistic program candidates of our interest.

## 2.1 Related work

Our work on learning probabilistic programs is related both to automatic programming and generalising from data. The former, automatic programming, addresses the synthesis of program code from specifications. Those specifications are often incomplete and include input/output examples. The latter, generalising from data, is one of the main aims of machine learning in as a whole.

### 2.1.1 Automatic programming

One recent overview of automatic programming is presented in (Gulwani et al., 2014) and its references. Approaches to program synthesis include work in inductive logic programming (Muggleton, 1996; Kersting, 2005; Raedt et al., 2008; Lin et al., 2014), evolutionary programming (Koza, 1992), inference over grammars (Olsson, 1995), and functional programming (Schmid and Wysotzki, 1998). From a very abstract standpoint, automatic programming concerns the search in the complex space of program code for programs that satisfy a certain specification. Automatic programming approaches differ from each other in the way that the specifications are formulated, in the expressiveness of the search space, and as to which search algorithm is used. Automatic programming is often paired with the field of programming languages and verification, in order to find programs which satisfy formal specifications.

Our approach is similar to the work on learning programs using a hierarchical Bayesian prior (Liang et al., 2010). In that paper authors also use a statistical inference framework, define a prior over program text and perform statistical inference. While they search for deterministic programs that satisfy several training input/output pairs, we look for probabilistic programs that are statistically similar to the distributions of in-terest. To the best of our knowledge, our work has been the first attempt to perform inference over probabilistic models in the form of probabilistic programs in such expressive probabilistic programming languages as Church, Venture or Anglican.

### 2.1.2 Generalising from data and automated modelling

The approach we describe in this chapter is related to density estimation (Silverman, 1986), which concerns the estimation of an unobservable probability density function given some observed data. There is, however, a major difference between density estimation and our approach. While most density estimation methods produce just a set of parameters (e.g. weights), we learn the representation of observed data in a structural and potentially interpretable form of generative model program code.

Our approach is also related to probabilistic model learning, for example to learning probabilistic relational models (Friedman et al., 1999) and Bayesian network structure (Mansinghka et al., 2012). Also worthwhile of mentioning are recent works in search over generative probabilistic model structures (Grosse et al., 2012) and kernel compositions (Duvenaud et al., 2013). Similarly to our approach, they explore a huge complex space of models, in which enumeration is intractable. While they use a greedy search algorithms to find an optimum model, we employ a fully Bayesian approach and define a non-parametric prior distribution over program text itself. This allows us to search over a more expressive class of probabilistic models, and allows us to penalise long or atypical program text.

## 2.2 Approach

We are interested in finding probabilistic programs, which when iteratively interpreted produce samples statistically similar to the distribution of interest $F_\lambda$ with parameter vector $\lambda$. For now, we assume that parameter vector $\lambda$ is fixed, and omit it to simplify notation. Each probabilistic program, which we consider as a potential match, is represented as its program text $\mathcal{T}$. We define the grammar prior over program text $p(\mathcal{T})$, details of which will be described in Section 2.4. The distribution of interest $F$ may be given in different forms; for example, as a set of samples $\mathcal{X} = \{x_i \sim F\}$, or as characteristics of that distribution $F$ (e.g. its moments).

For every particular program candidate $\mathcal{T}$, we want to evaluate how well it matches the distribution of interest $F$. There are no general ways to check this analytically by looking at its program text. We, therefore, employ the methods of approximate Bayesian computation. Specifically, we draw $N$ samples $\hat{\mathcal{X}} = (\hat{x}_1, \ldots, \hat{x}_J)$ from program $\mathcal{T}$ by evaluating it. (That is, $\hat{\mathcal{X}}$ is drawn from the distribution $p(\hat{\mathcal{X}}|\mathcal{T})$, since every probabilistic program text defines a distribution.) Before describing further details of our approach, we provide a brief outline of approximate Bayesian computation that is based on (Marin et al., 2012) and uses algorithms and equations there contained[1].

### 2.2.1 Basics of approximate Bayesian computation (ABC)

Let us start by considering the standard setup for Bayesian inference. There is some parameter of interest $\theta$, an intermediate hidden variable $\xi$, and an observation $y$. We are able to model the prior distribution $p(\theta)$, the condition distribution $p(\xi|\theta)$ for the intermediate hidden variable, and to sample observations from $p(\cdot|\xi)$. We are interested in the posterior

$$p(\theta|y) = \int p(\theta, \xi|y) d\xi \propto \int p(\theta, \xi) p(y|\theta, \xi) d\xi = \int p(\theta) p(\xi|\theta) p(y|\xi) d\xi.$$

For simplicity, let us agree to consider the intermediate variable $\xi$ to be the part of $\theta$, since we are able to marginalise over the part of variable $\theta := [\xi, \theta]$. This means that we have the prior distribution $p(\theta)$, the distribution $p(\cdot|\theta)$, and we are able to sample from both of them.

[1] In addition, notes (Huggins, 2013) from Jonathan Huggins were quite helpful.

Approximate Bayesian computation (ABC) framework (Marin et al., 2012) is different from the standard setup of Bayesian inference due to the impossibility (or intractability) to precisely calculate the likelihood $p(y|\theta)$, even if we can sample from $p(\cdot|\theta)$. The ABC approach proposes different algorithms to address this issue.

In the case when the distribution $p(\cdot|\theta)$ is finite or countable, one solution to obtain $N$ samples $\{\theta_i\}_{i=1}^N$ from $p(\theta|y)$ is to use the rejection sampling algorithm. This algorithm (Rubin et al., 1984) may be considered to be the first ABC algorithm, and it is presented.

**Algorithm 1** Likelihood-free rejection sampler (Marin et al., 2012)

```
for i = 1 to N do
    repeat
        Generate θ from the prior distribution p(θ)
        Generate z from the distribution p(·|θ)
    until z = y
    set θ_i = θ′,
end for
```

This algorithm cannot deal with cases where the sample space $\mathcal{D}$ of the distribution $p(\cdot|\theta)$ is neither finite nor countable. Algorithm 1 may be extended to the case of continuous sample spaces. While we cannot precisely compare $z$ and $y$, we may measure how similar they are. To do this, we need to introduce a statistic $\eta$ on the space of observations $\mathcal{D}$, introduce a metric $\rho$ on $\eta(\mathcal{D})$, and accept a new proposed $z$ if the distance between $\eta(y)$ and $\eta(z)$ is lower than the introduced fixed threshold $\epsilon$. This new algorithm is similar to the previous, and is presented below.

**Algorithm 2** Likelihood-free approximate rejection sampler (Marin et al., 2012)

**for** $i = 1$ to $N$ **do**
  **repeat**
    Generate $\theta$ from the prior distribution $p(\theta)$
    Generate $z$ from the distribution $p(\cdot|\theta)$
  **until** $\rho\{\eta(z), \eta(y)\} \leq \epsilon$
  set $\theta_i = \theta'$, $z_i = z'$,
**end for**

The above algorithm samples from the joint distribution (Marin et al., 2012)

$$p_\epsilon(\theta, z|y) = \frac{p(\theta)p(z|\theta)\mathbb{I}_{A_{\epsilon,y}}(z)}{\int_{A_{\epsilon,y}\times\theta} p(\theta)p(z|\theta)\,\mathrm{d}z\,d\theta}\,, \tag{2.1}$$

where $\mathbb{I}_A(\cdot)$ is the indicator function of the set $A$, and

$$A_{\epsilon,y} = \{z \in \mathcal{D} \mid \rho\{\eta(z), \eta(y)\} \leq \epsilon\}\,.$$

By marginalising this joint distribution 2.1 over $z$, we are able to get the sought after posterior distribution

$$p_\epsilon(\theta|y) = \int p_\epsilon(\theta, z|y)\,\mathrm{d}z \approx p(\theta \mid y)\,.$$

Finally, let us emphasise that there are three levels of approximation in such ABC algorithm:

1. the choice of statistics $\eta(\mathcal{D})$, for observations $y \in \mathcal{D}$, that often is chosen to be insufficient,
2. the distance function $\rho$ on $\eta(\mathcal{D})$,
3. and the value of tolerance level $\epsilon$.

### 2.2.2 Matching distributions using ABC

The previous subsection described the basics of ABC. Before describing a few more aspects of ABC, which are relevant for our work, let us begin to frame our problem in terms of ABC. Recall that in our set-up we have a distribution of interest $F$, which is given in a form of its characteristics, and we need to find the distribution $\mathcal{T}$ in the form of a probabilistic program, which matches $F$. We approach this problem in Bayesian way by providing a prior over samplers $p(\mathcal{T})$, and condition it to find such samplers $\mathcal{T}_1, \mathcal{T}_2, \ldots$ that match $F$.

Ideally, we would like to have a general way to compare two distributions, one of which is given as probabilistic program code and another is given by its characteristics. One would imagine doing this by drawing infinite number of samples from $\mathcal{T}$ or by some sort of code analysis, but this is, to the best of our knowledge, not feasible in the general case. As mentioned before, we do not know any better general way to check if program code $\mathcal{T}$ matches the distribution of interest $F$ rather than draw some number of samples from $\mathcal{T}$ and compare them to $F$. Therein lies a departure from the standard approximate Bayesian computation framework, which is that the type of ABC observation $\mathbf{y}$ is *distribution* in comparison to the type of *one-dimensional or multi-dimensional continuous variables* in common ABC settings. To represent and be able to compare this distribution to $F$, we need to sample from $\mathcal{T}$ some finite number of times $M$ to get $\hat{\mathcal{X}} = \{\hat{x}_1, \ldots, \hat{x}_M\}$ such that $\hat{x}_i \sim p(\cdot|\mathcal{T})$. Finally, we would like to marginalise over $\hat{\mathcal{X}}$ to approximate $p(\mathcal{T}|F)$.

One way to interpret this is to say that the whole statistic function $\eta(\cdot, M)$ is a composition of several helper statistic functions such that $\eta(\cdot, M) = \eta_1(\eta_2(\cdot, M))$. The fact that we draw only $M$ samples $\hat{\mathcal{X}}$ from $p(\cdot|\mathcal{T})$ is related to the partial contribution of statistic function $\eta_2(\cdot, M)$. In our case, the statistic $\eta_2(\cdot, M)$ will always be insufficient, since we cannot draw infinite number of samples. Thus, the whole statistic function $\eta$ will be definitely insufficient as well.

Once we have drawn $M$ samples from the distribution candidate $p(\cdot|\mathcal{T})$ of a probabilistic program candidate $\mathcal{T}$, we can use any statistical method to compare how similar it is to the distribution of interest $F$. This step is related to the contribution of another statistic function $\eta_1(\cdot)$. One example of such statistic is some finite number of raw and central moments. Here we would compare moments of $p(\cdot|\mathcal{T})$ and $F$ given the threshold $\epsilon$, so we accept only such $\mathcal{T}$ that $\rho(\eta_1(\hat{\mathcal{X}}), \eta_1(F))$, where $\hat{\mathcal{X}} = \{\hat{x}_i \sim p(\cdot|\mathcal{T})\}_{i=1}^{M}$. Details and examples are provided in further sections.

Let us summarise how our problem relates to ABC approach: the ABC prior distribution $p(\theta)$ is our prior over samplers $p(\mathcal{T})$, the ABC observation $z$ is the distribution $p(\cdot|\mathcal{T})$ that is defined by probabilistic program code $\mathcal{T}$, the "intermediate" statistic $\eta_2(z, M)$ of the distribution $z$ is the finite number of samples $\{x_i\}_{i=1}^{M}$ from $p(\cdot|\mathcal{T})$, the "full" statistic $\eta(z) = \eta(z, M) = \eta_1(\eta_2(z, M))$ is the statistic of the distribution $z$, the observation $y$ is the distribution of interest $F$, and the statistic $\eta(y)$ of the observation $y$ is the statistic $\eta(F)$ of the distribution of interest $F$. We consider that $y \approx z$ if the distance between two statistics $\rho(\eta(z), \eta(y)) < \epsilon$. Finally, our inference is intended to target an approximation of $p(\mathcal{T}|F)$, namely

$$p_\epsilon(\mathcal{T}|F) = \int p_\epsilon(\mathcal{T}, \hat{\mathcal{X}}|F) d\hat{\mathcal{X}} \propto \int p(\mathcal{T}) p(\hat{\mathcal{X}}|\mathcal{T}) \mathbb{I}_{A_{\epsilon,F}}(\hat{\mathcal{X}}) d\hat{\mathcal{X}} \,. \tag{2.2}$$

In the next section we describe how we can avoid using a fixed threshold $\epsilon$, and instead employ continuous distance functions that measure the similarity between distributions.

## 2.3 Noisy ABC

Another concept in ABC, that is essential for our work, is the proposal of "noisy ABC", made by Wilkinson in (Wilkinson, 2013). He suggests to replace the fixed threshold $\epsilon$ by a general noise kernel function $K_\epsilon(y, z)$, which should be a valid probability density function given a particular observation $y$. The kernel might be interpreted as measurement or model error. It is expected to have high values when $y \approx z$, and low values otherwise. The joint ABC target distribution becomes as follows (Marin et al., 2012):

$$p_{K,\epsilon}(\theta, z \mid y) = \frac{p(\theta)p(z \mid \theta)K_\epsilon(y, z)}{\int p(\theta)p(z \mid \theta)K_\epsilon(y, z)\, \mathrm{d}z\, \mathrm{d}\theta}\,, \tag{2.3}$$

$$p_{K,\epsilon}(\theta \mid y) \propto \int p(\theta)p(z \mid \theta)K_\epsilon(y, z)\, \mathrm{d}z\,. \tag{2.4}$$

The important point, which was made by Wilkinson himself in (Wilkinson, 2013), is that if the model already includes the error as part of its generative model (i.e. $K_\epsilon(y, z)$ is the part of the model in the same way as $p(\theta)$ and $p(z \mid \theta)$ are the parts of it), then the ABC will target the exact posterior for that model. If the kernel is approximate or added on the top of the model, then the ABC method will produce an approximate posterior estimate in general. In both cases, we are able to incorporate the kernel into probabilistic program code of the generative model and employ automatic general-purpose inference methods available in probabilistic programming engines. In particular, we used particle Markov chain Monte Carlo (PMCMC) inference in probabilistic program system Anglican (Tolpin et al., 2015b).

In the context of our work, the noisy ABC target $p(\mathcal{T}|F)$ is intended to be proportional to the marginal in $\hat{\mathcal{X}}$ of the joint distribution

$$p(\mathcal{T})p(\hat{\mathcal{X}}|T)\pi(F|\hat{\mathcal{X}})\,, \tag{2.5}$$

where $\pi(F|\hat{\mathcal{X}})$ is a kernel function that in our case measures a distance between two distributions. In other words, this is a penalty function that judges how distinct the samples $\hat{\mathcal{X}}$ from the program candidate $\mathcal{T}$ statistically are, as the whole, from the distribution of interest $F$.

### 2.3.1 Moments matching as an example of statistic $\eta_1$

Previously, we did not specify an example of statistic $\eta_1$ and kept it abstract. One example of such statistic is the set of moments of a distribution. The $n$-th moment, about a fixed value $c$, of the one-dimensional probability distribution, with existing probability density function $f$, is known to be:

$$\mu_n = \int_{-\infty}^{\infty} (x - c)^n \, f(x) \, \mathrm{d}x \,, \tag{2.6}$$

where the constant $c$ might be equal, for example, to zero (e.g. for the mean, which is the first raw moment) or to the mean (e.g. for the variance, which is the second central moment).

Moments are widely examined in statistics and have been used for a long time to estimate distribution parameters. Ideally, we would like to match infinitely many moments so as to precisely match at least some bounded distributions[2]. Unfortunately, in practice, we are limited, and can check only finite number of moments. This means that the statistic $\eta_1$ will also be insufficient. Up to some extent, we can control the precision of our ABC inference by increasing number of moments that we consider.

Figure 2.1 provides pseudocode of the probabilistic program that samples a probabilistic program `prog` (i.e. $\mathcal{T}$) from a grammar prior `grammar` (i.e. $p(\mathcal{T})$). There we

[2]In general, distributions cannot be uniquely identified even by the infinite number of their moments. One of counterexamples may be found in (Durrett, 2010).

```
(defquery lpp-normal
  (let
    [noise-level 0.001
     N 100
     prog-candidate-tuple (grammar `() `real)
     prog-candidate (extract-compound
                          prog-candidate-tuple)
     samples (apply-n-times prog-candidate N '())]
    (observe (normal (mean samples) noise-level) 0.0)
    (observe (normal (std  samples) noise-level) 1.0)
    (observe (normal (skew samples) noise-level) 0.0)
    (observe (normal (kurt samples) noise-level) 0.0)
    (predict (extract-text prog-candidate-tuple))))
```

Figure 2.1: Probabilistic program to infer program text for a $\text{Normal}(0, 1)$ sampler. Variable `noise-level` is a noise level. Procedure `grammar` samples a probabilistic program candidate from the adaptor-based grammar. This procedure `grammar` returns a tuple: a generated program candidate in the form of nested compound procedures (to draw samples $\hat{\mathcal{X}}$ and check them against the distribution of interest $F$), as well as the same program candidate in the form of program text to be interpreted by humans and re-used in the future.

To check how well the distribution of interest is matched by the probabilistic program candidate, we sample `N` samples from the latter. Then, we check how well those samples $\hat{\mathcal{X}}$ match the distribution of interest $F$, which, in this particular case, is specified by four moments.

aim to find probabilistic programs $\mathcal{T}$-$s$ that define distributions $\hat{x} \sim p(\cdot|\mathcal{T})$ with particular moments. Namely, the distribution $\hat{x} \sim p(\cdot|\mathcal{T})$ of sought $\mathcal{T}$ should have a mean of zero, a standard deviation of one, a skewness of zero, and an excess kurtosis of zero as well. As previously discussed, we approximate the distribution $p(\cdot|\mathcal{T})$ by drawing a finite number of `samples` (i.e. $\hat{\mathcal{X}}$) from $p(\cdot|\mathcal{T})$.

By seeking probabilistic programs that produce samples from distributions with those moments, we attempt to find such programs that draw samples from a distribution that is statistically similar to the standard Normal distribution. As discussed earlier, there are several levels of approximation, as we draw only finite number of samples from the program candidate $\mathcal{T}$, and as we also constrain only a finite number of moments. The way we constrain moments with the Gaussian noise kernel is also approximate.

### 2.3.2 Use of hypothesis test statistics for $\eta_1$

Another possible choice of statistic $\eta_1$ is a hypothesis test statistic. Figure 2.2 shows pseudocode that aims to find a sampler for the Bernoulli distribution family parametrised by $\lambda$ with the help of G-test statistic (McDonald, 2009):

$$G_{n,\lambda} = 2 \sum_{i \in 0,1} \#[\hat{\mathcal{X}}_n = i] \ln \left( \frac{\#[\hat{\mathcal{X}}_n = i]}{\lambda^i (1-\lambda)^{(1-i)} \cdot |\hat{\mathcal{X}}_n|} \right),$$

where $\#[\hat{\mathcal{X}}_n = i]$ is the number of samples in $\hat{\mathcal{X}}_n$ that take value $i$. We calculate G-test's p-value[3] of falsely rejecting a null hypothesis $H_0 : \hat{\mathcal{X}} \sim \text{Bernoulli}(\lambda_n)$. This p-value is incorporated to the kernel $\pi(F|\hat{\mathcal{X}})$ by observing a coin that is flipped and happens to face up with the probability equal to the p-value of the test.

As with the previous pseudocode for learning a standard Normal distribution sampler, in this example, we sample a program candidate from the grammar and draw $N$

[3]Which is not necessarily the probability of incorrectly rejecting the null hypothesis.

```
(defquery lpp-bernoulli
  (let
    [prog-candidate-tuple (grammar `(real) `int)
     proc-candidate (extract-compound
                         prog-candidate-tuple)
     N 100]
    (let
      [samples-1 (apply-n-times proc-candidate N '(0.5))]
      (observe (flip (G-test-p-value samples-1 `Bernoulli
                                     (list 0.5))) true))
    ...
    (let
      [samples-N (apply-n-times proc-candidate N '(0.7))]
      (observe (flip (G-test-p-value samples-N `Bernoulli
                                     (list 0.7))) true))
    (predict (extract-text prog-candidate-tuple))
    (predict (apply-n-times proc-candidate N '(0.3)))))
```

Figure 2.2: Probabilistic program to infer program text for a $\text{Bernoulli}(\lambda)$ sampler and generate `N` samples from the resulting procedure at a novel input argument value, $\lambda = 0.3$. Similar code was used to infer the program that samples from $\text{Poisson}(\lambda)$, which is a distribution with the infinite countable support. Procedure `(G-test-p-value samples target-dist target-params)` returns the p-value of falsely rejecting a hypothesis that `samples` are from the target distribution `target-dist` with fixed parameters `target-params`.

samples $\hat{\mathcal{X}}$ from the program candidate. Here, our task is more complex since we aim to find a sampler for the whole family of the Bernoulli distribution that is parametrised by one-dimensional parameter $\lambda \in (0, 1)$. This brings us to another important generalisation of our approach: we want to learn probabilistic programs, which are parametrised by inputs $\lambda$ and which define conditional distribution samplers. We may incorporate this into our ABC target, such that it becomes as follows:

$$p(\lambda)p(\mathcal{T}|\lambda)p(\hat{\mathcal{X}}|\mathcal{T}, \lambda)\pi(F|\hat{\mathcal{X}}, \lambda), \tag{2.7}$$

where $p(\lambda)$ is some prior over parameters. We are free to define the latter prior as we wish. We might construct that prior such that it is concentrated on regions of the parameter $\lambda$ that are of particular interest to us.

## 2.4 Prior over program code

We used grammar prior that is similar to one that was introduced in our preceding work (Perov and Wood, 2014; Perov, 2014). It complements the adaptor grammar (Johnson et al., 2007) prior that is used in (Liang et al., 2010) by the use of local environments[4] and type signatures.

The basic element in functional languages is an expression. To generate a probabilistic program we recursively apply the production rules listed below starting with $expr_{type}$, where $type$ is the desired output signature of the inducing program. Production rules are applied stochastically with some probabilities $\sum p_i = 1$ (covered in more detail later in this Section). The set of types used for our experiments is {`real`,

[4]An environment is a mapping of typed symbols to values (including such values as primitive and compound procedures), but these values are not evaluated/applied until the actual run of a probabilistic program. A compound procedure consists of formal arguments (just names) and its body, which is an expression to be evaluated. For example, compound procedure `(fn [x y] (+ x y))` has two formal arguments `x` and `y`, as well as body `(+ x y)`.

`bool`}. We also employ type `int`, which in our experiments were the derivative type of `real` with values rounded.

To avoid numerical errors while interpreting generated programs we replace functions like `log(a)` with `safe-log(a)`, which returns $0$ if $a < 0$, and `uniform-continuous` with `safe-uc(a, b)` which swaps arguments if $a > b$ and returns $a$ if $a = b$. The general set of procedures in the global environment include `+, -, *, safe-div, safe-uc, cos, safe-sqrt, safe-log, exp, inc, dec`.

An example of the production rules code, written in Anglican, is provided in Figure 2.3. Schematically our prior is defined below:

1. $expr_{type} \mid env \xrightarrow{p_1} v$,

   where variable $v$ is a randomly chosen variable from the environment $env$ such that it has type $type$. An example of a sampled program using this rule:

   ```
   (fn [my-var another-var] my-var).
   ```

2. $expr_{type} \mid env \xrightarrow{p_2} c$,

   where $c$ is a random constant $c$ with the type $type$. Constants were drawn from the predefined constants set (including $0.0$, $\pi$, etc.) and from normal and uniform continuous distributions. Example:

   ```
   (fn [my-var] 0.3).
   ```

3. $expr_{type} \mid env \xrightarrow{p_3} (procedure_{type}\ expr_{arg_1_type} \ldots expr_{arg_N_type})$,

   where $procedure$ is a primitive or compound, and deterministic or stochastic procedure, which is chosen randomly from the global environment with output type signature $type$. Examples:

   ```
   (fn [my-var] (safe-uc ... ...)).
   ```

```
(fn [my-var] (+ ... ...)).
```

4. $expr_{type} \mid env \xrightarrow{p5} (let\ [new\text{-}symbol\ expr_{any}]\ expr_{type} \mid env\ \cup\ new\text{-}symbol))$, where $env\ \cup\ new\text{-}symbol$ is an extended environment with a new variable named $new\text{-}symbol$ (each time a symbol name is unique, e.g. generated by Lisp's `gensym`). The value of the new variable is defined by an expression, which is generated according to the same production rules. The variable has fixed but arbitrary type (i.e. type "any"), which is chosen randomly from the set of employed types (i.e. `real`, `bool` or `int`). Examples:

```
(fn [my-var] (let [x (safe-uc -1 1)] (+ x x))).

(fn [my-var] (let [x (cos (safe-uc -1 1))] ...)).

(fn [my-var] (let [x (safe-uc -1 1)] (+ x my-var))).
```

5. $expr_{type} \mid env \xrightarrow{p4}$
$(let\ [new\text{-}symbol\ (fn\ \ formal\text{-}args\ \ expr_{any} \mid env\ \cup\ formal\text{-}args\}]$
$expr_{type} \mid env\ \cup\ new\text{-}symbol)$,
where `formal-args` is the list of unique, previously not used, symbol names. The body of the compound procedure is generated using the same production rules, given an environment that incorporates variables `formal-args`. After the compound procedure is defined in the environment, it might be used in the body of the let. Possible general examples:

```
(fn [my-var]
  (let
    [cp1 (fn [x] (* x 2.3))]
    (cp1 (cos my-var))))
```

```
(fn [my-var]
  (let
    [cp1 (fn [x] (* x 2.3))]
    (cp1 (cp1 my-var))))
```

6. $expr_{type} \mid env \xrightarrow{p_6} (if\ (expr_{bool})\ expr_{type}\ expr_{type})$.

7. $expr_{type} \mid env \xrightarrow{p_7} (recur\ expr_{arg_1_type}\ \ldots\ expr_{arg_M_type})$,

   i.e. recursive call to the current compound procedure if we are inside it, or to the main inducing procedure, otherwise. Possible example:

```
(fn [val]
  (if (= val 1)
    1
    (* val (recur (- val 1)))))
```

```
(def productions
  (fn productions [expression-type provided-args-names env-variables-names
                   procedure-name recursion-is-okay depth-level
                   max-allowed-depth-level cur-pos]
    (cond
      (= expression-type 'real)
        (let
          [expression-option
            (if (> depth-level (min max-allowed-depth-level
                    MAX-ALLOWED-DEPTH-FOR-PRODUCTION-GRAMMAR))
              ; If we reached the max depth level,
              ; we want just a constant or a variable lookup:
              (sample
               distribution-on-real-expressions-only-constants-and-local-variables)
              (sample distribution-on-real-expressions))]
          (cond
            (= expression-option 0)
              (let [for-body (get-real-constant)]
                (list for-body (fn [provided-args env-variables] for-body)))
            (= expression-option 2)
              (let
                [chosen-procedure-id (sample distribution-on-binary-real-operators)
                 chosen-procedure-name
                   (nth (list '+ '- '* 'safe-div 'safe-uc) chosen-procedure-id)
                 chosen-procedure
                   (nth (list + - * safe-div safe-uc) chosen-procedure-id)
                 new-max-allowed-depth-level
                   (if recursion-is-okay
                     (+ depth-level (sample-depth-max-increase))
                     max-allowed-depth-level)
                 chosen-operand1-mix
                   (productions 'real provided-args-names env-variables-names
                                procedure-name false (inc depth-level)
                                new-max-allowed-depth-level 'arg)
                 chosen-operand2-mix
                   (productions 'real provided-args-names env-variables-names
                                procedure-name false (inc depth-level)
                                new-max-allowed-depth-level 'arg)
                 chosen-operand1 (second chosen-operand1-mix)
                 chosen-operand2 (second chosen-operand2-mix)]
                (list
                  (list chosen-procedure-name (first chosen-operand1-mix)
                        (first chosen-operand2-mix))
                  (fn [provided-args env-variables]
                    (chosen-procedure
                      (chosen-operand1 provided-args env-variables)
                      (chosen-operand2 provided-args env-variables)))))
            ...))
      ...)))
```

Figure 2.3: Production rules as a probabilistic program in Anglican. The procedure returns a tuple of the generated program candidate, and its human interpretable program text. The program candidate is generated in the form of nested compound procedures.

Prior probabilities $\{p_i\}$ were automatically extracted from a small corpus of sampler source code written in Anglican language. The corpus was manually prepared and was based on one-dimensional distribution sampler code from (Devroye, 1986; Box and Muller, 1958; Knuth, 1998). The corpus is provided in Appendix A. The prior was employed in our experiments in a manner similar to cross-validation. For example, when we were learning a sampler for the standard Normal distribution, we held out the source code for the standard Normal distribution and the general Normal distribution from the corpus. In addition, our prior was smoothed by Dirichlet priors.

To ensure that program candidates terminate, we allow only 10 nested self-recursive calls. If that limit is reached, a procedure deterministically returns $0.0$.

Figure 2.4 illustrates how probable some of the sought probabilistic programs are given our production rules. Figure 2.5 illustrates the flexibility of our prior over code for one-dimensional samplers. It shows samples from some random probabilistic programs, which were sampled from the grammar prior.

## 2.5 Experiments

The initial experiments had been described in our prior work (Perov, 2014; Perov and Wood, 2014). In the current work we show in Section 2.5.1 that our approach is comparable to evolutionary algorithms, a common method for program synthesis. Then we re-implement our method in new probabilistic programming systems, namely *Anglican* and *Probabilistic Scheme*. This provided us with a ten-fold improvement in speed, as reported in Section 2.5.2. Finally, in Section 2.5.3, we report new experimental results, similar to (Perov, 2014; Perov and Wood, 2014), but with thinner binning.

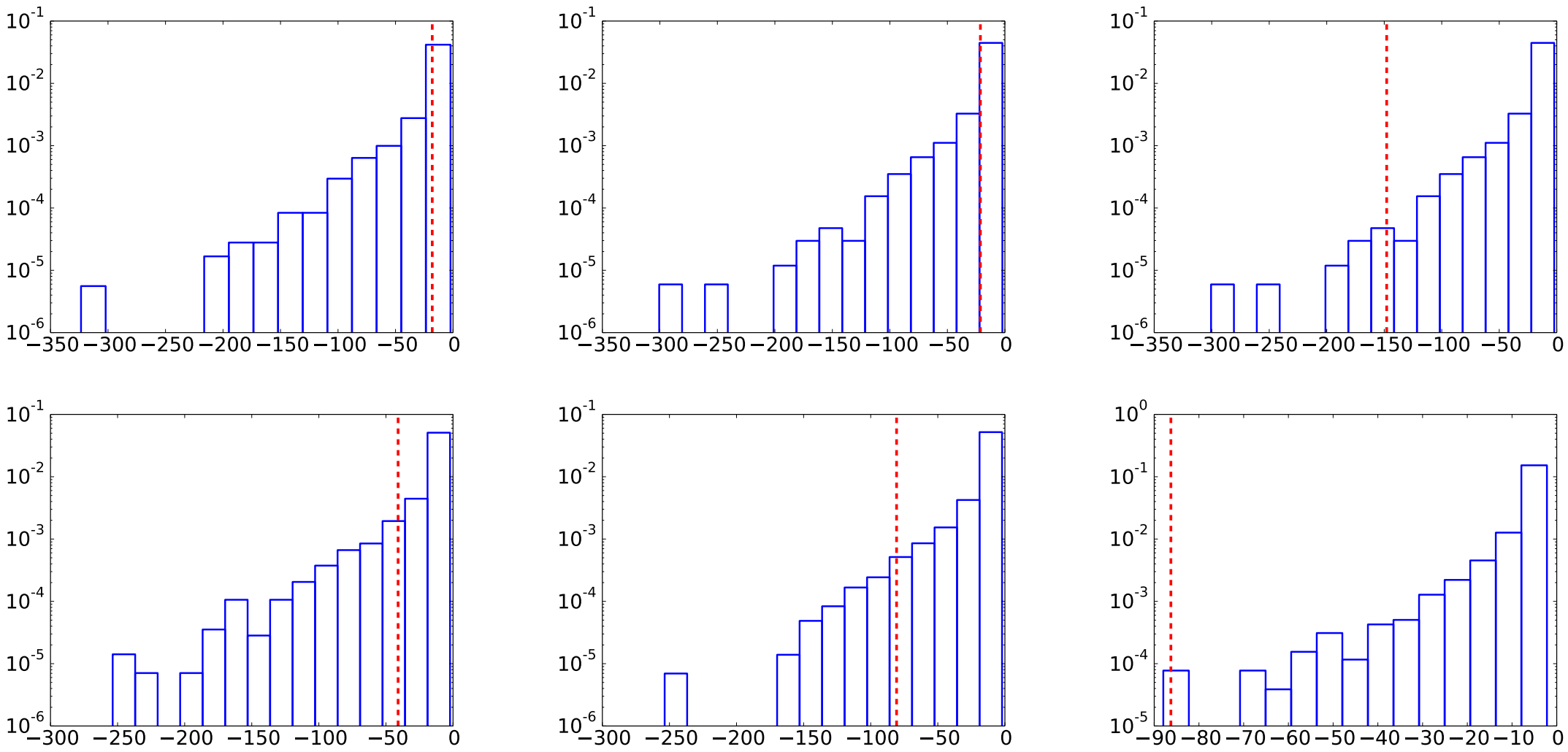

Figure 2.4: Blue histograms illustrate frequencies of log-probability of probabilistic programs, which are sampled from the grammar prior $P(\mathcal{T})$ that is used in the search of random variate generators for: *(top row, left to right)* $\text{Bernoulli}(p)$, $\text{Beta}(a, 1.0)$, $\text{Gamma}(a, 1.0)$, *(bottom row, same)* $\text{Geometric}(p)$, $\text{Poisson}(\lambda)$, $\text{Normal}(0, 1)$. Red dashed lines show the log probabilities, given the production rules, of program code for samplers that is written by humans and manually translated to Anglican code.

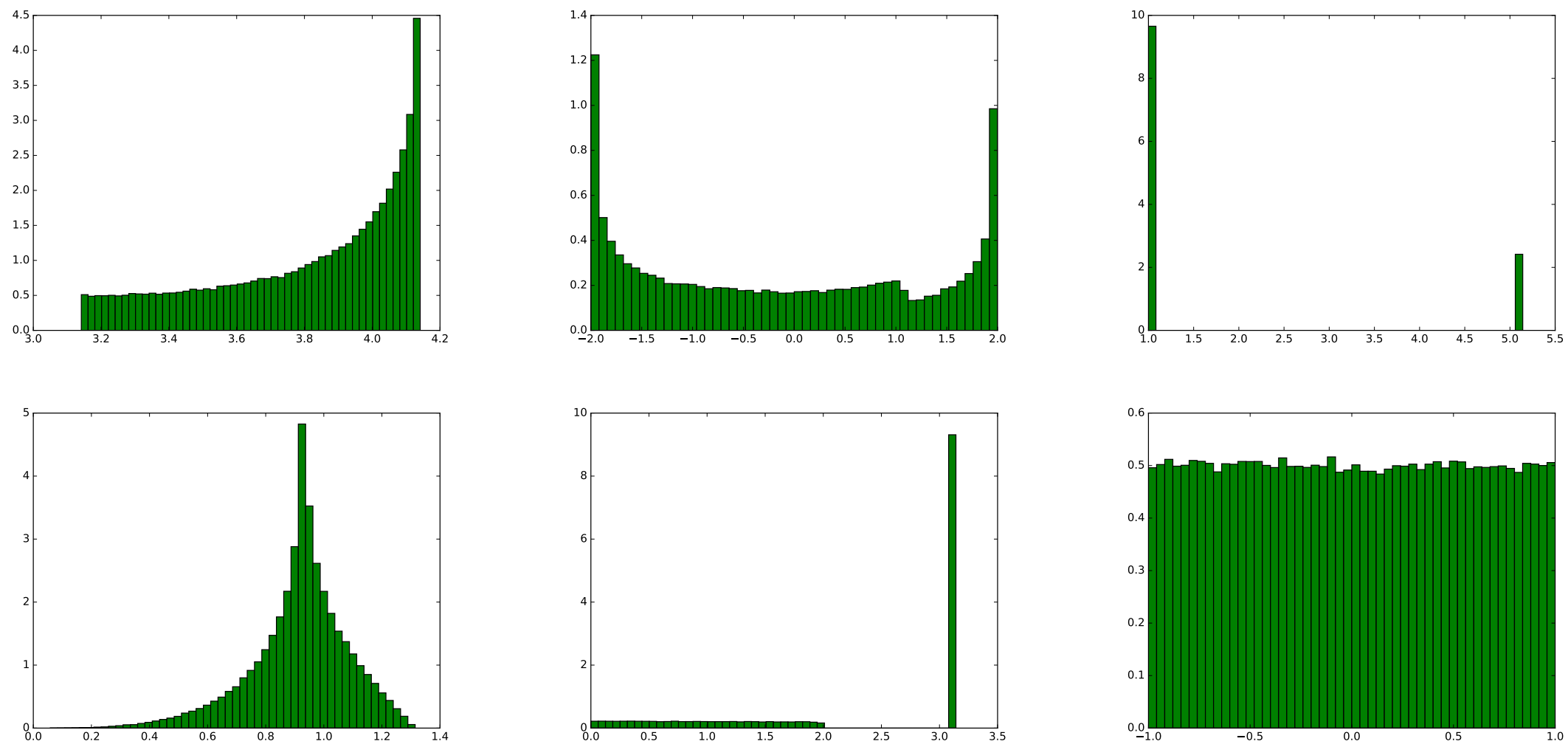

Figure 2.5: Normalised histograms of 200,000 i.i.d. samples from 6 different samplers, code for which is randomly unconditionally generated from our adaptor-grammar-like prior `grammar`.

### 2.5.1 Evaluation of our approach versus evolutionary algorithms

Our approach was evaluated against genetic programming (Koza, 1992), one of state-of-the-art methods to search in the space of programs. Genetic programming is an evolutionary based metaheuristic optimisation algorithm that is used to generate a program given the specification. For a recent introduction into the field of genetic programming, see (Poli et al., 2008). The very similar grammar, which we described in Section 2.4, was reproduced in the evolutionary computation framework DEAP (Fortin et al., 2012) written in Python. The fitness function was selected as the log probability presented in the Equation 2.5 with the $p\left(\hat{X} \mid \mathcal{T}\right)$ term omitted, in accordance with the assumption that sought probabilistic programs will repeatedly appear in the results of search over programs. An alternative would be to marginalise over $\hat{X}$. However, that requires more program runs and is, therefore, more computationally expensive.

We used the DEAP framework to generate individual (program) code. Then, "individuals" (i.e. program code candidates in genetic programming vocabulary) were evaluated in Anglican. We decided to use Anglican for this, and not Python, because functional-style `let` construction is not naturally supported in Python. Finally, evaluation results (log probabilities) were reported back to DEAP to generate and select individuals for the next generation. We had 100 individuals per generation and used DEAP set-up for the strong typed genetic programming optimisation.

Figure 2.6 shows that PMCMC inference performance is similar to genetic programming. In contrast to genetic programming, PMCMC is a statistically valid estimator of the target distribution. In addition, the probabilistic programming system allows reasoning about the model over models and the inference of models within the same framework, while genetic programming is an external machinery which considers the optimisation of probabilistic programs in a black box way.

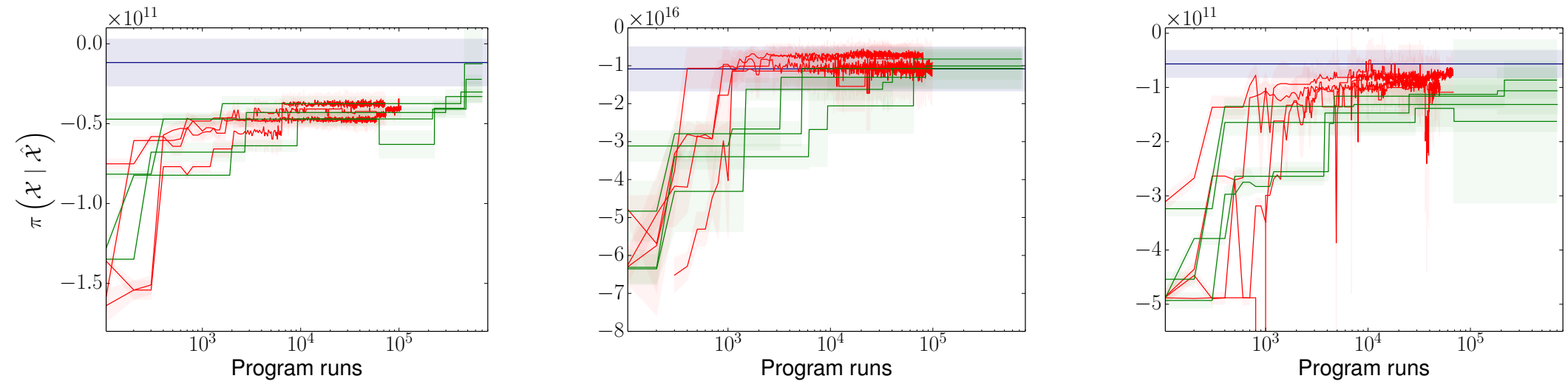

Figure 2.6: Convergence of unnormalised penalty function $\pi\left(\mathcal{X} \mid \hat{\mathcal{X}}\right)$ for $\mathrm{Bernoulli}(p)$, $\mathrm{Normal}(\mu, \sigma)$, and $\mathrm{Geometric}(p)$ correspondingly. $\hat{\mathcal{X}}$ is a samples set from a probabilistic program $\mathcal{T}$ as described in Section 2.2. Navy lines show the true sampler's penalty function value (averaged by 30 trials), red lines correspondent to genetic programming, and green lines – to PMCMC. Transparent filled intervals represent standard deviations within trials. We ran a smaller number of genetic programming runs because their evaluation took more time in our set-up.

### 2.5.2 Engines comparison

Figure 2.7 shows a speed comparison between different probabilistic programming engines: *Anglican* (Tolpin et al., 2015b)[5], *Interpreted Anglican* (Wood et al., 2014)[6] and *Probabilistic Scheme* (Paige and Wood, 2014). Our prior work (Perov and Wood, 2014; Perov, 2014) had been done in *Interpreted Anglican* (Wood et al., 2014).

*Interpreted Anglican* engine is written in Clojure and interprets Anglican code. *Anglican* engine is written in and integrated with Clojure. It treats control structures and translates them to Clojure code. Thus, Clojure is a compilation target for *Anglican*. *Probabilistic Scheme* engine is based on Scheme compiler "Stalin" [7], which is written in C, with included Probabilistic C (Paige and Wood, 2014) library. Probabilistic program Scheme code is thus a compilation target for Probabilistic Scheme compiler that is enhanced by Probabilistic C.

[5] https://bitbucket.org/probprog/anglican
[6] https://bitbucket.org/probprog/interpreted-anglican
[7] https://en.wikipedia.org/wiki/Stalin_(Scheme_implementation)

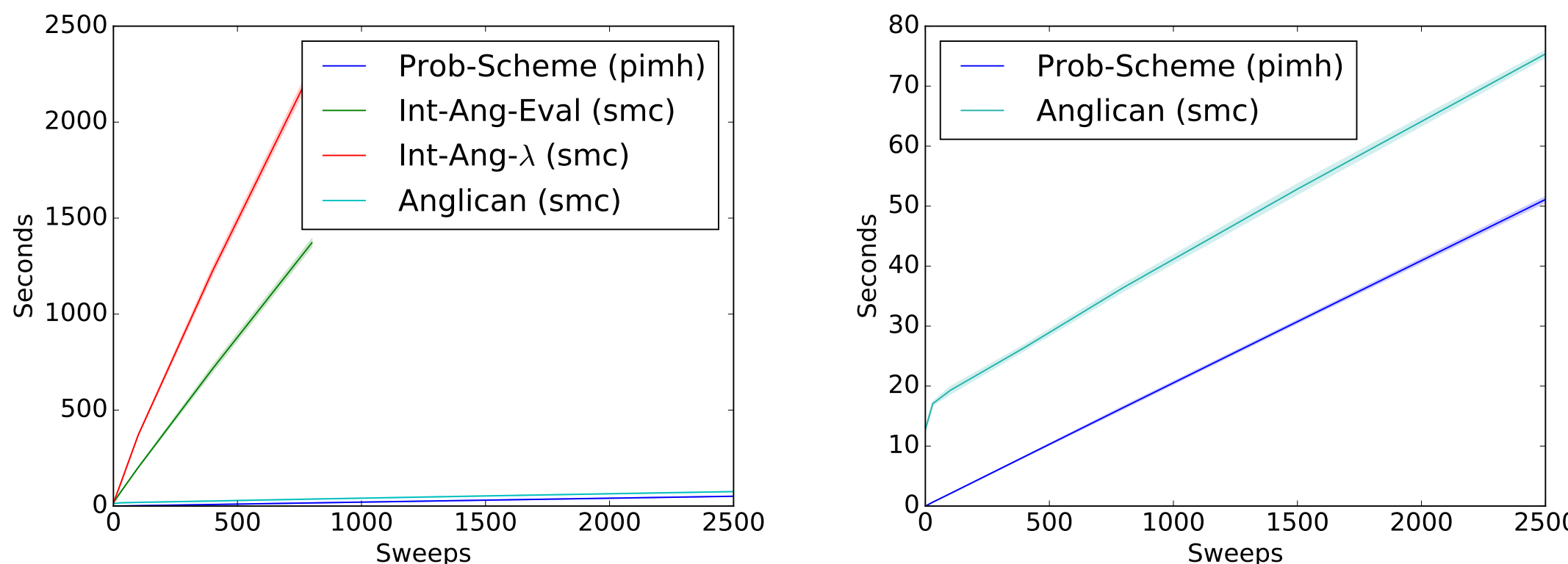

Figure 2.7: Performance comparison for different probabilistic programming engines: Anglican, `Anglican`; Interpreted Anglican, `Int-Ang`; Probabilistic Scheme, `Prob-Scheme`.

*Interpreted Anglican* is the only amongst those to support `eval`. There is thus an additional graph “Int-Ang-Eval (smc)” that shows the time complexity for not generating nested compound procedures, but instead generating full Anglican program text and then evaluating this text to a compound procedure. We prefer the former approach of generating nested compound procedures, since it is supported in more probabilistic programming systems and it provides an opportunity for inference optimisations in the future. For example, one might think of doing local proposals on the internal sub-body of the probabilistic program candidate body. An illustration of such potential proposal is that the following program

```
(fn [...] ... (+ a (sin b)) ...)
```

might be changed, with the probability of the proposal, according to the inference kernel (e.g., MH or Gibbs), according to the generative model and given the observed distribution of interest, to

```
(fn [...] ... (- b a) ...).
```

New probabilistic programming systems, *Anglican* and *Probabilistic Scheme*, made the process of learning probabilistic programs at least 10 times faster. This is a significant constant-factor improvement, especially taking into the consideration the fact that the space over probabilistic program code is complex. In addition, the approach runs faster in Probabilistic Scheme, most probably because it is compiled to C++. For our latest experiments we decided to employ *Anglican*, as it has been easier for us to develop and especially debug our experiments in Clojure.

### 2.5.3 Learning sampler code

Given the improvement in speed performance, we were able to reproduce the initial experiments much faster and report results with better accuracy. In particular, in Figure 2.8 we show results of learning sampler program code for six common one-dimensional distributions $\mathrm{Bernoulli}(p)$, $\mathrm{Poisson}(\lambda)$, $\mathrm{Gamma}(a, 1.0)$, $\mathrm{Beta}(a, 1)$, $\mathrm{Normal}(0, 1)$, $\mathrm{Normal}(\mu, \sigma)$. As before, we marginalised over the parameter space with a small randomly composed set of $\lambda_1, \ldots, \lambda_S$. Figure 2.9 shows repeated experiments for learning independent one-dimensional samplers that aim to match arbitrary one-dimensional real world empirical data from a credit approval dataset[8] (Quinlan, 1987; Bache and Lichman, 2013).

[8]We used continuous fields A2, A3, and A8. There were a bit more than 650 data points for each dimension. See `https://archive.ics.uci.edu/ml/machine-learning-databases/credit-screening/crx.names` for details.

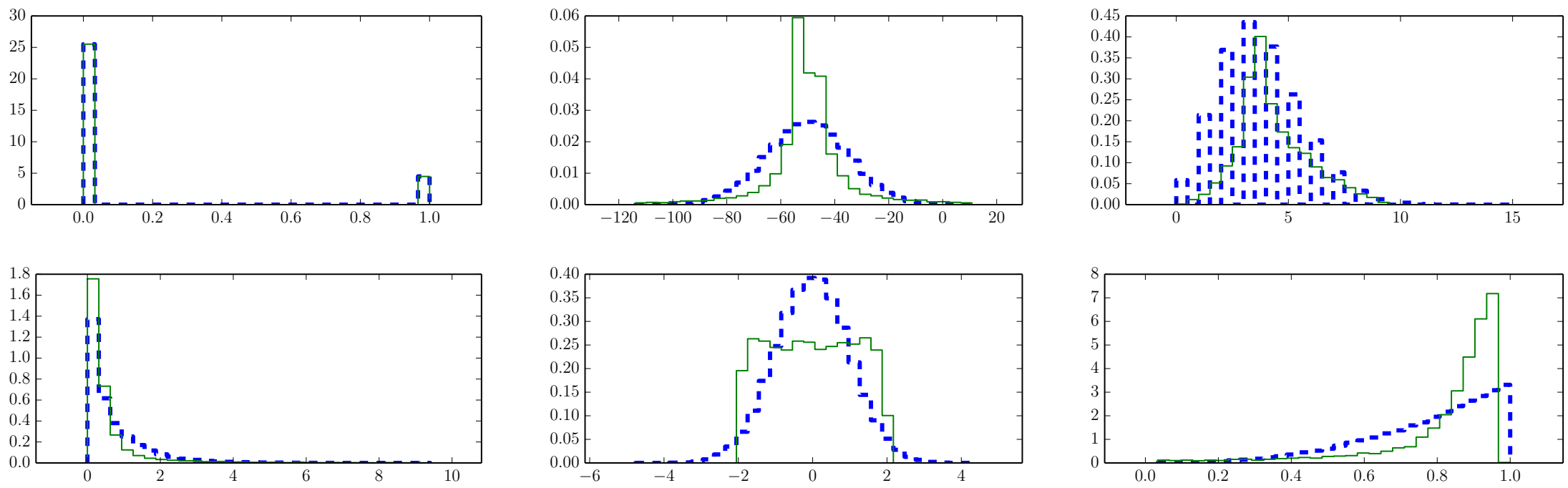

Figure 2.8: Representative histograms of samples *(green solid lines)* drawn by repeatedly evaluating one of the best learnt sampler program text versus *(blue dashed lines)* histograms of exact samples drawn from the corresponding true distribution. Top row, left to right: $\mathrm{Bernoulli}(p)$, $\mathrm{Normal}(\mu, \sigma)$, $\mathrm{Poisson}(\lambda)$. Bottom row, left to right: $\mathrm{Gamma}(a, 1.0)$, $\mathrm{Normal}(0, 1)$, $\mathrm{Beta}(a, 1.0)$. The parameters used to produce these plots do not appear in the training data. In the case of $\mathrm{Bernoulli}(p)$ we again were able to infer the program code that exactly matches the whole family of the Bernoulli distribution. Not all finite-time inference converges to good approximate sampler code as illustrated by some examples.

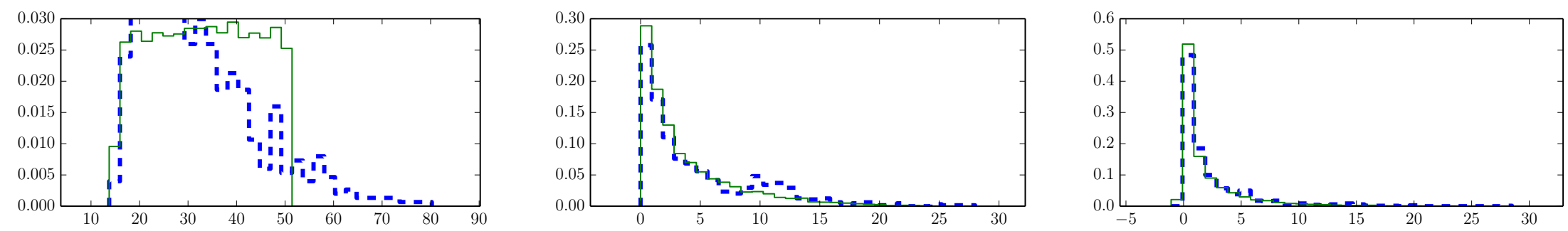

Figure 2.9: Histograms of samples *(green solid lines)* generated by repeatedly interpreting inferred sampler program text and the empirical distributions *(blue dashed)* which were trained to match data from a real world dataset.

### 2.5.4 Example of learning a standard Normal sampler

We illustrate the process of learning a sampler for the standard Normal distribution by providing a few examples of probabilistic program text from the Markov chain that targets the posterior for such sampler. In Figure 2.10 we provide samples that are generated by evaluating those probabilistic programs. This Section provides program text for each corresponding subplot in order: left to right, top to bottom. One of the first inferred program, from the posterior over program text, is a program that always deterministically returns 0.0. This very short program text has a very high probability given the production rules, and it precisely matches the first moment, i.e. the mean:

```
1 (lambda (stack_level) 0.0)
```

Next sensible approximations, from the Markov chain, are probabilistic programs that sample from an uniform continuous distribution with fixed bounds:

```
1 (lambda (stack_level) (safe-uc -1.0 1.0))
```

```
1 (lambda (stack_level) (safe-uc -2.0 (+ 3.14159 -1.0)))
```

```
1 (lambda (stack_level)
2   (begin
3     (define G__3352 -1.0)
4     (safe-uc G__3352 (+ 1.0 (safe-div 0.0 0.0)))))
```

Finally, the chain converges to more complex and more precise approximations to a standard Normal distribution sampler:

```
1 (lambda (stack_level)
2   (+ (safe-uc (+ -2.0 3.14159) (inc -1.0)) (safe-uc 1.0 -2.0)))
```

```
(lambda (stack_level)
  (* (begin (define G__56510
    (safe-div (safe-uc 1.0 (begin (define G__56511
      (safe-uc 1.0 (safe-log 0.0))) G__56511)) 1.0))
        (safe-uc -2.0 (exp G__56510)))
          (safe-sqrt (safe-uc 0.0 1.0))))
```

```
(lambda (stack_level)
  (safe-uc (safe-uc 0.0 (exp 1.0)) (* -1.0 1.0)))
```

```
(lambda (stack_level)
  (* (begin (define G__56510 (safe-div
    (safe-uc 1.0 (begin (define G__56511
      (safe-uc 1.0 (safe-log 0.0))) G__56511)) 1.0))
        (safe-uc -2.0 (exp G__56510)))
          (safe-sqrt (safe-uc 0.0 1.0))))
```

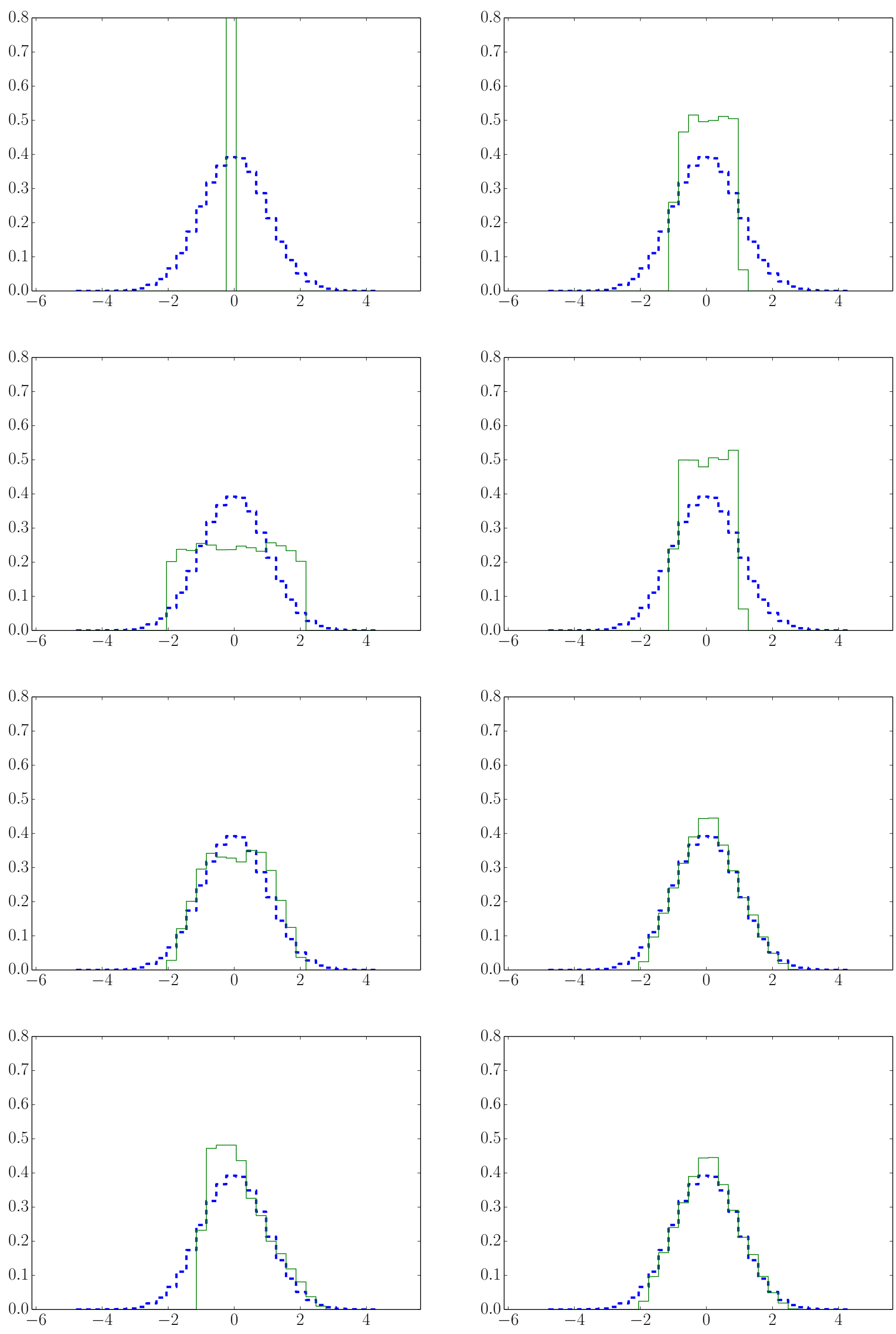

Figure 2.10: Different samples of program code, from the Markov chain (particle Markov Chain Monte Carlo inference in Anglican), that aim to statistically match the sampler for the standard Normal distribution. See source code in Section 2.5.4.

## 2.6 Outline of a motivating example for the scientific rediscovery in classical genetics

Our approach allows us to automatically infer generative models. One of motivations for such inference is to induce laws of nature. As an example, we outline a set of experiments that aim to scientific rediscovery of laws of classical genetics. These laws might be induced in the form of probabilistic programs. The proposed experiments are designed to follow Mendel's original experiments (Mendel, 1985). In this thesis we only outline the potential set of experiments, thus leaving the experiments themselves as potential future work by ourselves or by others.

### 2.6.1 Aims of the proposing experiment

We aim to induce three laws of classical genetics in the form of probabilistic programs (Wikipedia, 2015):

1. Law of Segregation: (a) "an individual contains a pair of alleles for each particular trait which segregate or separate during cell division for this trait", (b) and "each parent passes a randomly selected copy (allele) to its offspring".

2. Law of Dominance: "there is a notion of dominant and recessive alleles, and a recessive allele is always masked by a dominant allele".

3. Law of Independent Assortment: "separate genes for separate traits are passed independently of one another from parents to offspring".

### 2.6.2 Brief introduction to Mendel's work

Mendel designed and conducted his experiments in plant hybridisation, aiming to find "a generally applicable law governing the formation and development of hybrids". He presented his results in his classic paper "Experiments in Plant Hybridisation" (1865) (Mendel, 1985). His well-known paper contains most of his experimental results in aggregated form that we can use as observations.

The parts of his work, which might be relevant for the possible scientific rediscovery of classical genetics laws, can be briefly described as follows[9]:

1. He carefully considered requirements on what plants to use. In Section 2 he describes that he chose the species *Leguminosae*, in particular the genus *Pisum*.

2. He selected 22 varieties of this genus. A variety is "a group of organisms that are members of the same species, and have certain characteristics in common that are not shared by all members of the species". Plants from different varieties are differentiable by several characteristics, but they can still be hybridised with each other since they relate to the same genus.

3. He first cultivated these varieties independently in order to verify that within the variety, plants remain constant without exceptions. This is quite important, since this is also the part of experiments that would need to be reproduced. He does not provide any data for this part of his experiments, and thus it is necessary to guess as to how many plants and for how many generations he cultivated varieties independently.

4. In Section 3 he defines 7 pairs of strongly differentiating characteristics between variates. He decides to consider these differentiating characteristics as binary (e.g. the difference in the form of the ripe pods is strongly distinguished to be "either simply inflated, not contracted in places; or they are deeply constricted between the seeds and more or less wrinkled"). He supports this decision by the evidence that there is no transition forms both in original varieties and in hybrids.

[9] To the best of my understanding.

5. For each pair, he fertilises varieties with different observable characteristics, with each other. "Each of the two varieties which in one set of fertilizations served as seed-bearer in the other set was used as the pollen plant." In this way he produces hybrids.

6. In Section 4 he elaborates on the forms of hybrids which he obtained. He defines the terms "dominant" and "recessive" characteristics, by noting that the recessive characteristics always "withdraw or entirely disappear in the hybrids, but nevertheless reappear unchanged in their progeny". For each pair of characteristics he identifies what characteristic is the dominant of the two. He does not provide much data for this part of his experiments, and thus it is again necessary to guess how many hybrids he examined for each of trait.

7. In Section 5 he reports on a first generation derived from hybrids (so it is the second generation from original varieties). He reports the average proportion of 3:1 for the dominant and recessive characteristics respectively. He provides aggregate data for each pair of characteristics.

8. In Section 6 he reports that "those forms which in the first generation exhibit the recessive characteristic [i.e. one third] do not further vary in the second generation as regards this characteristic; they remain constant in their offspring". "Of these [other] two-thirds yield offspring which display the dominant and recessive characteristics in the proportion of 3:1, . . . ".

9. In Section 7 he reports that the subsequent generations from the hybrids follow

the same rule as that discovered in Sections 5 and 6. In Section 8 he conducts experiments on 2 or 3 different characteristics together. He reports that "the relation of each pair of different characteristics in hybrid union is independent of other differences in the two original parental stocks". In Section 9 he examines the reproductive cells of the hybrids.

### 2.6.3 Abstractions to induce

The aim of the experiment is to induce three probabilistic procedures that represent (a) an abstraction of a plant, (b) an abstraction of an observable feature of a plant, and (c) an abstraction of plants hybridisation. Below we elaborate on them, and briefly describe how these abstractions together can be interpreted as the three laws of Mendelian genetics.

### 2.6.4 Inducing a probabilistic procedure

For the purpose of inducing probabilistic program code we refer to an abstract procedure `sample-probabilistic-procedure` that randomly generates a probabilistic program by sampling from some grammar over program text, which is similar to the one described in Section 2.4. This procedure takes two arguments:

1. The list of input signatures of the inducing probabilistic procedure.
2. The output signature of the inducing probabilistic procedure.

For example, if we want to induce program code of a probabilistic procedure which takes three real values and which returns the list of two lists both containing the pair of integers, we need to call the procedure `sample-probabilistic-procedure` as follows:

```
(define new-program
  (sample-probabilistic-procedure
    (list 'real 'real 'real)
    (list (list 'int 'int) (list 'int 'int))))
```

Below are provided two possible outcomes for `new-program` sampled from `sample-probabilistic-procedure`:

```
(lambda (a b c)
  (list
    (list a (+ a b))
    (list b (safe/ a c))))
```

```
(lambda (a b c)
  (list
    (list (normal b c) (+ a a))
    (list (safe-uniform-continuous a b) 3)))
```

This is similar to the procedure `grammar` in Section 2.3.1.

### 2.6.5 An abstraction of an individual plant

Firstly, we assume that an individual plant is just some data. In other words, the assumption is that to describe a real natural object it is enough to represent this object in the form of finite data. We need to induce a produce `produce-random-individual` that produces a random individual plant (like a factory). This procedure takes no arguments and returns an unknown, but some specific data structure. It is specific since we assume that all our plants and their offspring belong to a single species, even though they may belong to different varieties of this species:

```
(define produce-random-individual
  (induce-probabilistic-procedure
    '()  ; No inputs.
    'any ; Any, but fixed output signature.
  ))
```

By calling the procedure `produce-random-individual` we sample individual plants:

```
(define plant-1 (produce-random-individual))
```

```
(define plant-2 (produce-random-individual))
```

By an abstraction of a plant, we mean its entire life cycle, including time spent as a seed before it actually grows and becomes a real plant.

### 2.6.6 An abstraction of an observable plant feature

Secondly, we need to have a probabilistic procedure `get-feature` that, based on an individual plant's data, samples an observable feature. This procedure should take an individual, and return a feature. Following Mendel's work, we work with strongly distinguishable binary features (like the colour is only purple or white). This means that we can assume that the output signature of the inducing procedure to be boolean:

```
(define individual-data-signature
  (get-output-signature produce-random-individual))
(define get-feature
  (induce-probabilistic-procedure
    (list individual-data-signature) ; One input:
                                     ; the individual data.
    'bool ; The output is boolean.
          ; We rely on the fact that we work with
          ; a highly differentiable characteristic which
          ; does not have any gradient:
          ; it is just one or another.
  ))
```

The procedure `get-output-signature` is just a language helper function that returns the output signature of any procedure.

### 2.6.7 An abstraction of the plant hybridisation process

Thirdly, we need to have a probabilistic procedure which hybridises plants, such that a new plant is obtained. We aim to induce two fully independent probabilistic procedures: the procedure `produce-hybrid-sexually` and the procedure `produce-hybrid-asexually`:

Procedure `produce-hybrid-sexually` takes as an input two individual plants, and produces a new plant:

```
(define produce-hybrid-sexually
  (induce-probabilistic-procedure
    (list (output-signature produce-random-individual)
          (output-signature produce-random-individual))
    (output-signature produce-random-individual)))
```

Procedure `produce-hybrid-sexually` takes as input an individual plant, and produces a new one:

```
\begin{lstlisting}
(define produce-hybrid-asexually
  (induce-probabilistic-procedure
    (list (output-signature produce-random-individual))
    (output-signature produce-random-individual)))
```

### 2.6.8 Procedures that are expected to be learnt

Using our current understanding of classic genetics, in Figures 2.11, 2.12, 2.13 and 2.14 we provide below the source code for probabilistic programs which could be expected to induce the basic laws of classical genetics. That is, we provide source code for four probabilistic programs: `produce-random-individual`, `get-feature`, `produce-hybrid-sexually`, `produce-hybrid-asexually`. For simplicity, we deal with only one trait (genetically determined characteristic).

```
(define produce-random-individual
  (lambda ()
    (cons (flip)
      (cons (flip)
        nil))))
```

Figure 2.11: Interpretation: "each individual contains two alleles for each trait".

```
(define get-feature
  (lambda (me)
    (if (car me) True (second me))))
    ;; This is the implementation
    ;; of logical OR.
```

Figure 2.12: Interpretation: Law of Dominance states that "recessive alleles will always be masked by dominant alleles". Procedure `second` is just `(lambda (x) (car (cdr x)))`.

```
(define produce-hybrid-sexually
  (lambda (pollen-parent egg-parent)
    (cons (if (flip) (first pollen-parent)
                     (second pollen-parent))
      (cons (if (flip) (first egg-parent)
                       (second egg-parent))
        nil))))
```

Figure 2.13: Interpretation: The Law of Segregation states that "a pair of alleles of an individual for each particular trait is being segregated during cell division (assuming diploidy) for any particular trait", and "that each parent passes a randomly selected copy (allele) to its offspring". The function is symmetric.

```
(define produce-hybrid-asexually
  (lambda (parent)
    (produce-hybrid-sexually parent parent)))
```

Figure 2.14: Interpretation: "asexual hybridisation is similar to a sexual one".

To prove the Law of Independent Assortment (i.e. that "separate genes for separate traits are passed independently of one another from parents to offspring") we need to consider several traits at once in the same way as Mendel did in his experiments as described in Section 8 of his paper.

In the proposed experiments, the same order might be followed as in Mendel's work. This means that experiments are started for different traits independently.

The experiments that are proposed in this Section may constitute the future work in the direction of learning probabilistic programs.

## 2.7 Conclusion

In this Chapter we described an approach to learning samplers in the form of program code within the framework of probabilistic programming. The initial results have been encouraging for the future work on automatically learning problem-specific generative models given data and just a few examples.

Although we were able to reproduce the inference of the precise sampler code for the Bernoulli family distribution, we were only able to find approximate samplers for other one-dimesional families like the Normal or Gamma distributions. Better inference techniques and hierarchical/cumulative learning methods (Henderson, 2010; Dechter et al., 2013) are essential to finding more complex probabilistic programs, including human-interpretable programs as in Appendix A that are theoretically in the prior of our grammar but were not identified during our experiments. Better inference is especially important for the automatic induction of probabilistic programs that are internally conditioned on observations (i.e. to automatically induce problem-specific generative models that are similar to, for example, hidden Markov or latent Dirichlet allocation models; or even more sophisticated probabilistic problems that aim to describe our world and agents in it, for example as in (Stuhlmüller, 2015) and in (ForestDB, 2016)). Ultimately, it should be fruitful to add higher-order types into the grammar (like $function < int >$, $function < real >$, $function < * >$, where $*$ is a wild-card, in addition to $int$, $real$, $bool$, etc.), such that the grammar procedure is capable of producing itself (in other word, the grammar procedure code has a non-zero probability under the same grammar). By doing so, we will be able to receive an agent, in the form of a probabilistic program, that can both solve tasks and produce more advanced agents.

We appreciate as well that our results on finding one dimension representations of unknown distributions are not capable of outperforming the state-of-the-art methods for density estimation and related problems. The goal of the research described in this Chapter was to make an initial step towards learning probabilistic programs to describe potentially interpretable and composable generative models of the world.

We also outlined a set of experiments concerning classical genetics that may constitute future work on applying the approach of learning probabilistic programs for scientific rediscovery and, eventually, discovery.

In addition, while we briefly showed that within the set of basic experiments we have received similar results to the optimisation-based method of genetic programming, the real applications of the program induction and synthesis would most probably benefit from the combination of statistical methods, such as MCMC, and optimisation methods, such as evoluationary algorithms or stochastic search methods. Finally, inference over program code should also benefit from better proposals using discriminative models (e.g. as in (Karpathy et al., 2015), and potentially in combination with (Reed and de Freitas, 2016)), similar to ones that are described in the following Chapter.

# Chapter 3

# Data-driven proposals in probabilistic programming

As discussed in Chapter 1, to make probabilistic programming more efficient and widely used by the machine learning community, we need to enhance the quality and speed of statistical inference. This is also crucial for both existing and new applications, including our work on learning probabilistic programs. This section describes a way to facilitate sequential Monte Carlo inference with data-driven proposals using discriminative models.

### 3.0.1 Proposals in sequential Monte Carlo inference

In applications of sequential Monte Carlo inference, including probabilistic programming systems, the initial proposal distribution $q(\mathbf{x}_t | \mathbf{x}^s_{1:t-1}, y_{1:t})$ for the particle $s$ is often equal to the prior distributions $p(\mathbf{x}_t | \mathbf{x}^s_{1:t-1}, y_{1:t-1})$ of the generative model. This is known as a "generate and test" approach (Murphy, 2012) since we just sample values $\mathbf{x}_t$ from the generative model and only then evaluate how good they fit a data point $y_t$. Another name for this approach is "bootstrap particle filter" (Gordon et al., 1993). This approach does not require the specification of any additional parameters, and, therefore, it is convenient. The posterior is usually, however, different from the generative model prior, and thus the convergence of this simple "generate and test" approach can be quite slow.

It is important to note that sample spaces for $\mathbf{x}_t$, as well as corresponding probability measures for both conditional distributions $p(\mathbf{x}_t | \mathbf{x}^s_{1:t-1}, y_{1:t-1})$ and $q(\mathbf{x}_t | \mathbf{x}^s_{1:t-1}, y_{1:t})$, are not trivial because number of generated random variables in $\mathbf{x}_t$ is not fixed, and may even be unbounded. For example, consider a probabilistic procedure

```
(lambda (w) (if (sample (flip w)) 0 (+ 1 (recur))))
```

that samples from a geometric distribution with parameter $w$. In addition, even the type and structure of random variables in $\mathbf{x}_t$ might be not fixed. Because of that, in this work we consider only such proposal distributions $q$ that have the same structure of random choices as $p$ does.

To accelerate inference, we would like to take into the account the data $y_t$ so that

$$q(\mathbf{x}_t | \mathbf{x}^s_{1:t-1}, y_{1:t}) = p(\mathbf{x}_t | \mathbf{x}^s_{1:t-1}, y_{1:t}) = \frac{p(y_t | \mathbf{x}_{1:t}, y_{1:t-1}) p(\mathbf{x}_t | \mathbf{x}^s_{1:t-1}, y_{1:t-1})}{p(y_t | \mathbf{x}^s_{1:t-1}, y_{1:t-1})}.$$

This proposal is called "a fully-adapted proposal". When we use this proposal, the new weight

$$w^s_t \propto w^s_{t-1} p(y_t | \mathbf{x}^s_{1:t-1}, y_{1:t-1}) = w^s_{t-1} \int p(y_t | \mathbf{x}'_{1:t}, y_{1:t-1}) p(\mathbf{x}'_t | \mathbf{x}^s_{t-1}, y_{1:t-1}) d\mathbf{x}'_t. \quad (3.1)$$

This is the optimal proposal because for any given $\mathbf{x}^s_{t-1}$ the new weight $w^s_t$ will have the same value independently of the value of $\mathbf{x}^s_t$ (Murphy, 2012). This means that, conditional on the old values $\mathbf{x}^\cdot_{t-1}$, the variance of weights $w^\cdot_t$ is equal to zero.

### 3.0.2 Using a discriminative model for data-driven proposals

In general, it is intractable to calculate the integral in Equation 3.1 and therefore sample from $p(\mathbf{x}_t | \rho_t)$ directly, where $\rho_t$ is an environment $(\mathbf{x}^s_{1:t-1}, y_{1:t})$. However, we can

approximate this distribution with another distribution $\mathfrak{q}(\mathbf{x}_t|\eta)$, which is parametrised by unknown parameters $\eta(\rho_t)$. Such proposals are referred to as "data-driven". We use a discriminative model $\mathfrak{N}$ to map features of the environment $\phi(\rho_t)$ to parameters $\eta$. The aim of training the discriminative model is to bring $\mathfrak{q}(\mathbf{x}_t|\eta = \mathfrak{N}(\phi(\rho_t)))$ as close as possible to $p(\mathbf{x}_t|\rho_t)$. For now, let us also assume that $t$ is fixed and we are learning a proposal for some specific $\mathbf{x}_t$.

To train the discriminative model, we need $M$ pairs of training inputs $\{\phi(\rho_t)\}_j$ and related outputs $\{\mathbf{x}_t\}_j$ such that each $\mathbf{x}_t$ is drawn from the desired distribution $p(\mathbf{x}_t|\rho_t)$ using some "approximate" Monte Carlo sampling method, e.g. SMC or PMCMC. We call Monte Carlo methods "approximate" since we are never capable of having infinite number of particles in SMC or infinite number of MCMC iterations in PMCMC.

There are at least two approaches to obtaining these training pairs. The first is the unconditional offline simulation from the generative model, by sampling and capturing both latent variables $\mathbf{x}_t$ and observations $y_t$. The second approach is to run a thorough offline sequential Monte Carlo inference on some number of training episodes, and then to capture the values $\mathbf{x}_t$ from the estimated filtering distribution $\widetilde{p}(\mathbf{x}_t|\rho_t) = \sum_s w_s \delta_{\mathbf{x}_t^s}(\mathbf{x}_t^s)$.

Recall that our aim is to provide a good approximation to $p(\mathbf{x}_t|\rho_t)$ in the form of $\mathfrak{q}(\mathbf{x}_t|\eta)$, which is parametrised by the discriminative model output $\eta$. The choice of distribution type for $\mathfrak{q}(\mathbf{x}_t|\eta)$ is flexible. As mentioned, it should have the same structure of random choices as the prior $p(\mathbf{x}_t|\rho_t)$. In addition to being able to sample from it, we have to be able to calculate sample's normalised likelihood. As for the choice of a discriminative model, one option is to use feedforward neural networks:

$$\eta = \mathfrak{N}_\theta(\phi(\rho_t)),$$

where $\theta$ are parameters of the neural network to learn. Here the values of $\phi(\rho_t)$ constitute the input layer, and the parametrisation $\eta$ of $\mathfrak{q}(\mathbf{x}_t|\eta)$ constitutes the output layer of the neural network. Lastly, $\theta$ are parameters of the neural network we need to learn.

We look for such $\theta$ that $\mathfrak{q}(\mathbf{x}_t|\mathfrak{N}_\theta(\phi(\rho_t)))$ is close to $p(\mathbf{x}_t|\rho_t)$. Hence, the loss function $L_t(\theta)$ should depend on a measure of the difference between these two probability distributions. Specifically, we use the Kullback-Leibler divergence

$$D_{KL}(p(\mathbf{x}_t|\rho_t) \mid\mid \mathfrak{q}(\mathbf{x}_t|\mathfrak{N}_\theta(\phi(\rho_t)))) = \mathbb{E}_{p(\mathbf{x}_t|\rho_t)} \left[\log p(\mathbf{x}_t|\rho_t) - \log \mathfrak{q}(\mathbf{x}_t|\mathfrak{N}_\theta(\phi(\rho_t))))\right].$$

Let us highlight that by we need to maximise the similarity of the whole the family of distributions $\mathfrak{q}(\mathbf{x}_t|\mathfrak{N}_\theta(\phi(\rho_t)))$ to another family $p(\mathbf{x}_t|\rho_t)$, not just the similarity of one distribution to another. This means we have to take another expectation over parameters $\rho_t$. Therefore, the loss function $L_t(\theta)$ approximates two nested expected values:

$$\begin{aligned}
L_t(\theta) &\approx \mathbb{E}_{p(\rho_t)} \left[D_{KL}(p(\mathbf{x}_t|\rho_t) \mid\mid \mathfrak{q}(\mathbf{x}_t|\mathfrak{N}_\theta(\phi(\rho_t))))\right] \\
&= \mathbb{E}_{p(\rho_t)} \left[\mathbb{E}_{p(\mathbf{x}_t|\rho_t)} \left[\log p(\mathbf{x}_t|\rho_t) - \log \mathfrak{q}(\mathbf{x}_t|\mathfrak{N}_\theta(\phi(\rho_t))))\right]\right] \\
&= \mathbb{E}_{p(\rho_t,\mathbf{x}_t)} \left[\log p(\mathbf{x}_t|\rho_t) - \log \mathfrak{q}(\mathbf{x}_t|\mathfrak{N}_\theta(\phi(\rho_t))))\right].
\end{aligned}$$

Ignoring the first term since it is a constant w.r.t. $\theta$:

$$L_t(\theta) \approx -\mathbb{E}_{p(\rho_t,\mathbf{x}_t)} \left[\log \mathfrak{q}(\mathbf{x}_t|\mathfrak{N}_\theta(\phi(\rho_t))))\right].$$

In practice, we do not integrate over all possible observations, but instead consider a specific set of training observation values, such that

$$L_t(\theta) \approx -\mathbb{E}_{p(\rho_t,\mathbf{x}_t|y_{1:t})} \left[\log \mathfrak{q}(\mathbf{x}_t|\mathfrak{N}_\theta(\phi(\rho_t))))\right].$$

Once we replace the expectations by their approximation, the loss function for the discriminative model becomes as follows:

$$L_t(\theta) = -\sum_{j=1}^{M} w_{s(j)} \log \mathfrak{q}(\mathbf{x}_t = \mathbf{x}_t^{s(j)} \mid \eta = \mathfrak{N}_\theta(\phi(\mathbf{x}_{1:t-1}^{s(j)}, y_{1:t}))), \tag{3.2}$$

where values $\mathbf{x}_*^*$ and weights $w_*$ are obtained by exhaustive inference runs (e.g., SMC).

In many models $\mathbf{x}_t$ and $y_t$ are homogeneous. Therefore in such models we can learn one $\mathfrak{q}(\cdot|\eta)$ for all $t$. This means that $L(\theta) = \sum_t L_t(\theta)$.

Since our proposal $\mathfrak{q}(\mathbf{x}_t|\eta)$ will only be an approximation to $p(\mathbf{x}_t|z_{1:T}, y_{1:T})$, we do not set the weight $w_t^s \propto w_{t-1}^s p(y_t|\mathbf{x}_{t-1}^s)$. Instead, we calculate the full equation:

$$w_t^s \propto \frac{p(y_t|\mathbf{x}_{1:t-1}^s, y_{1:t-1})p(\mathbf{x}_t^s|\mathbf{x}_{1:t-1}^s, y_{1:t-1})}{\mathfrak{q}(\mathbf{x}_t|\eta)}.$$

This should not be a problem since we assume that we know the likelihood function of $\mathfrak{q}(\mathbf{x}_t|\eta)$.

Finally, in practice, we may use other distributions to approximate the filtering distribution $p(\mathbf{x}_t|\rho_t)$. For example, we may use the approximation of the smoothing distribution $p(\mathbf{x}_t|z_{1:T}, y_{1:T})$. This is as most of statistical inference in existing probabilistic programming systems is directed towards the approximation of the smoothing distribution.

### 3.0.3 Experiments with the linear Gaussian model

As a basic proof of concept, we test the approach on a linear Gaussian model with the following parameters:

$$\begin{aligned} x_0 &\sim \mathrm{Normal}(0, 2), \\ x_i &\sim \mathrm{Normal}(x_{i-1}, 0.1^2), \\ y_i &\sim \mathrm{Normal}(x_i, 0.1^2). \end{aligned}$$

For this experiment, identity features $\phi(\rho_t)$, provided as input to the neural network, are the values of last 10 latent states $x_{t-10}, \ldots, x_{t-1}$ and last 10 observed values of $y_{t-9}, \ldots, y_t$. The neural network has therefore 20 input nodes, with a 25 node hidden layer. The proposal distribution $\mathfrak{q}(x_t \mid \ldots) \sim \text{Normal}(x_t \mid (\mu, \sigma^2) = N_\theta(\phi(\rho_t)))$ is the Normal distribution with two unknown parameters $\mu$ and $\sigma$, and hence there are 2 nodes on the output layer, one for each of those.

We use the sigmoid function to connect the input layer to the hidden layer, and use the identity and exponential functions to connect the hidden layer with the output nodes $\mu$ and $\sigma$.

To train the neural network, we follow the second approach described above, thus producing several synthetic training and test episodes. By an episode we just mean here a dataset $\{y_1, \ldots, y_T\}$. While we run SMC inference with 100 particles on training episodes, we capture the values $x_t$ from the estimated smoothing distribution. Initial training episodes' observations and estimated latent variables will exactly constitute all the necessary training data for our neural network. Details on how we generate data for episodes are provided in the next subsection.

#### 3.0.3.1 Functions to generate training and test episodes $\{y_{1:T}\}$

To generate data for our training and test episodes we use following periodic functions:

1. Step training functions: `square(t)`, `square(t - pi/6)`, `square(t + pi/6)`,`square(t - pi/4)`,`square(t + pi/4)`,`square(t + pi/3)`, `square(t - pi/3)`,`square(t + pi/2)`,`square(t - pi/2)`. Function `square(t)` is a MATLAB square wave function[1] which is similar to `sin(t)`, but instead of a sine waves it creates peaks of $\pm 1$.

2. Smooth training functions: `sin(t - pi/6)`, `sin(t + pi/6)`, `sin(t - pi/4)`,`sin(t + pi/4)`,`sin(t - pi/3)`,`sin(t + pi/3)`,`sin(t -`

[1] See details by the link: `http://uk.mathworks.com/help/signal/ref/square.html`.

`pi/2),sin(t + pi/2).`

Test functions are similar to train functions, but have different offsets:

1. Step test functions: `square(t - 1),square(t + 1),square(t - 2), square(t + 2).`
2. Smooth test functions: `sin(t - 1), sin(t + 1), sin(t - 2), sin(t + 2).`

Each episode is a list of outputs of these functions with argument $t$ uniformly distributed in the range $[1, 100]$ with the step size of $\delta = 0.5$. The Gaussian noise $\mathrm{Normal}(0.0, 0.1^2)$ is added to the values. A subset of the training and test episodes is given in Figure 3.1 for illustration purposes.

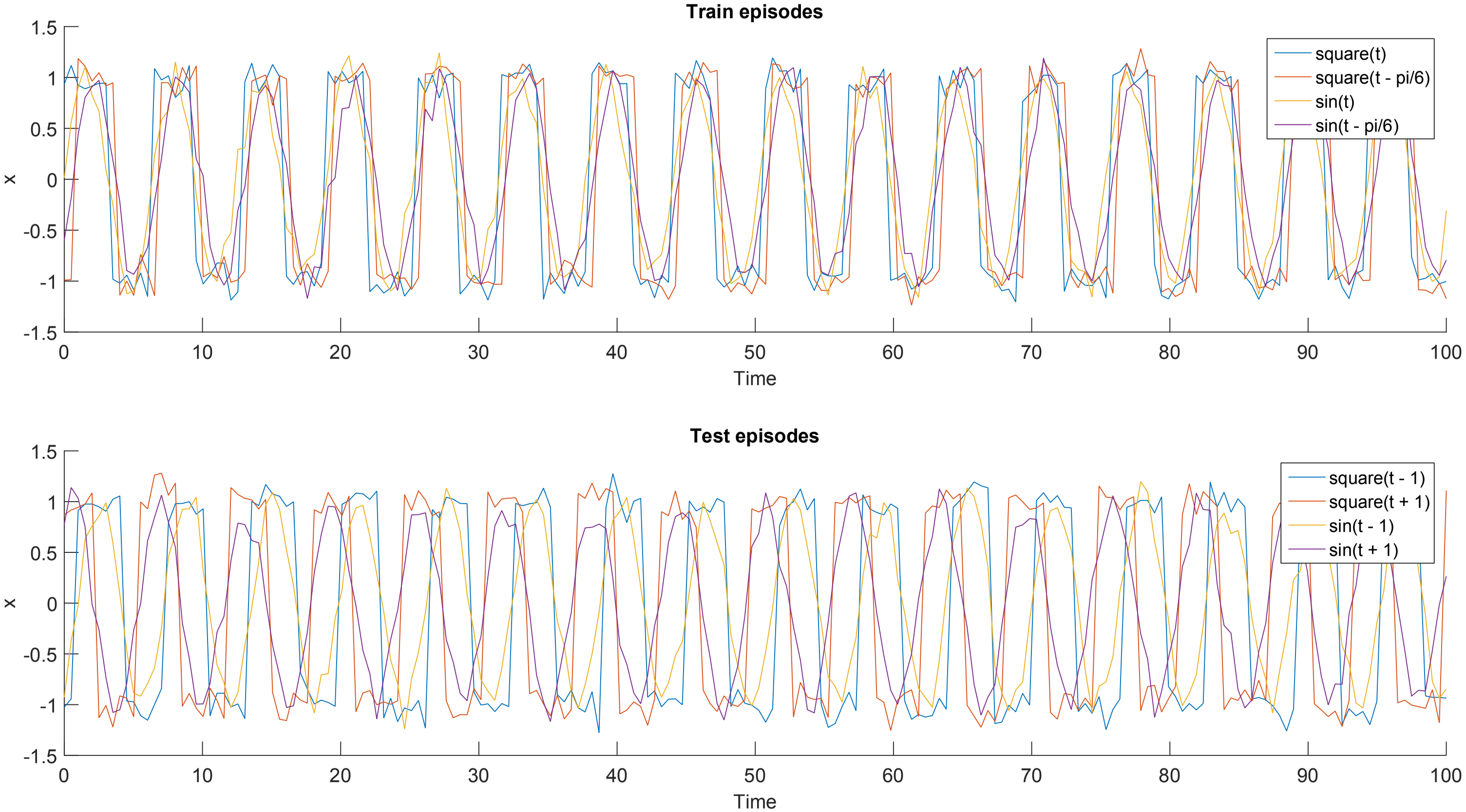

Figure 3.1: Subset of training and test functions.

#### 3.0.3.2 Comparing sequential Monte Carlo runs without and with data-driven proposals

Subfigures 3.2a and 3.2b show, on the left, average errors per latent state running SMC continuously with 100 particles on training episodes' data. By continuously, we mean that once SMC has been run on the episode, we additionally train the neural network on the data from the estimated smoothing distribution. To take account of all previous episodes, we incrementally add training data for the neural network after each episode. In addition, once SMC for the episode has been run and new training data has been added, we start neural network learning algorithm with parameters learnt after the previous episode. On their right, Subfigures 3.2a and 3.2b show average errors per latent state if we run SMC on test episodes with 10 particles with and without neural networks proposals. We run SMC for each episode independently. When we use neural network proposals, we use them with probability $p = 0.7$; that is, with probability $p = 0.3$ we still sample from the model prior. Experiments were repeated five times. Bars show standard deviations of average errors per latent state.

These basic experiments show that data-driven proposals with a disriminative model may significantly improve sequential Monte Carlo, especially when the prior distribution of the model is misspecified (the variance of $x_i \sim \mathrm{Normal}(x_{i-1}, 0.1^2)$ was purposefully quite low for the provided training and test data).

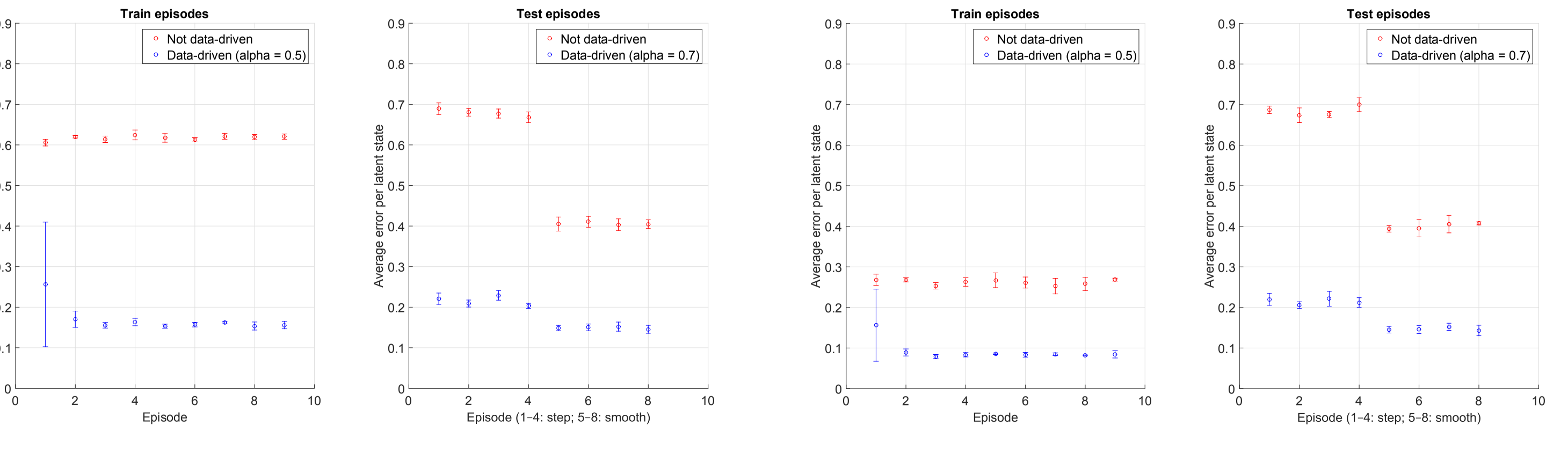

(a) Training the discriminative model only on step train functions *(left)*, and testing it on all test functions *(right)*.

(b) Training the discriminative model only on smooth train functions *(left)*, and testing it on all test functions. *(right)*

Figure 3.2: Average errors per latent state.

### 3.0.4 Applying the approach to the DDPMO model

Basic experiments with the linear Gaussian model have shown promising results. We would like to verify the approach on a more complex model. It would be also helpful to generalise our approach so that it may be easily applied to other models, which are written in one of probabilistic programming languages. For our further experiments we chose a dependent Dirichlet Process mixture of objects (DDPMO) model (Neiswanger et al., 2014). This is a recent Bayesian non-parametric model for detection-free tracking and object modeling. The model is based on a generalised Pólya urn (GPU) for time-varying Dirichlet process mixtures (Caron et al., 2007).

The DDPMO models the position and colour $\mathbf{y}_{t,n}$ of a foreground pixel $n$ in a video frame $t$ as an observed variable. This observed variable $\mathbf{y}_{t,n}$ depends on the latent variables of the model, such as cluster assignments $c_{t,1:N_t}$ and object parameters $\theta_t^k$ for each cluster $k$. The DDPMO is a native Bayesian non-parametric model, since the number of clusters and the related object parameters is unbounded and dependent on the observed data. The generative process of the DDPMO is described in (Neiswanger et al., 2014). The comparison of the object recognition and tracking performance of Bayesian statistical inference in the DDPMO model against the performance of some others state-of-the-art models and methods (not necessarily Bayesian) is also provided in (Neiswanger et al., 2014).

#### 3.0.4.1 DDPMO model in Anglican

We have expressed the DDPMO model as a 190 line Anglican program, and the GPU as another 75 line program. The GPU code may be reused in the future. See Appendix B for source code.

#### 3.0.4.2 Conjugate priors

The DDPMO model, as presented in (Neiswanger et al., 2014), uses conjugate priors. In particular, it uses a Multinomial distribution with a Dirichlet prior and a Multivariate normal distribution with a Normal-inverse-Wishart prior. Conjugate priors may be implemented in languages like Church (Goodman et al., 2008), Anglican (Wood et al., 2014) and Venture (Mansinghka et al., 2014) in the form of exchangeable random procedures (XRPs). Conjugate priors give a closed-form expression for the likelihood with marginalised-out prior distribution parameters, in this case Dirichlet and Normal-inverse-Wishart priors. While conjugate priors are not necessary, inference often becomes tractable only with the conjugate priors. That is because with conjugate priors we do not need to spend excessive computational resources integrating over hyperparameters in the process of Monte Carlo inference.

To perform inference in DDPMO, a library of additional XRPs for Anglican has been implemented, including procedure `mvn-conjugate-fast` for a Multinomial distribution with a Dirichlet prior, and `dirichlet-multinomial` for a Multivariate normal distribution with a Normal-inverse-Wishart prior. This library is also made such that it can be reused in the future.

#### 3.0.4.3 Data-driven proposal for SMC inference in DDPMO

We would like to obtain a better proposal for a cluster assignment of any new data point (pixel) in the DDPMO. Recall that the model is non-parametric, so that the number of clusters is not fixed. Below we provide a specification of the input and output of a discriminative model used as the proposal. It defines what data is needed from the current model state in order to perform a proposal on the cluster assignment for a new pixel.

**Inputs.** The features $\phi(\rho_t)$ of the environment $\rho_k = (\mathbf{x}_{t-1}, y_{1:t})$, which are the inputs to the neural network, consist of the following:

- Distances to the three already existing nearest clusters, of $K$, in the ascending order, $d_i \in \mathbb{R}, i = 1, \ldots, 3$.
- Colour histograms of a $7 \times 7$ patch surrounding these three clusters in the discretised HSV space, normalised to sum to one, such that $h_i \in \mathbb{R}^{10}, \sum_{j=1}^{10} h_{ij} = 1$.
- Colour histogram of a $7 \times 7$ patch surrounding the new data point (i.e. pixel) in the discretised HSV space, normalised to sum to one, $c \in \mathbb{R}^{10}, \sum_{i=1}^{10} c_i = 1$.

**Output.** We use the just described features $\phi(\rho_t)$, which reduce an undefined number of clusters to the three[2] closest ones and all others. This allows us to aim to approximate the posterior of a discrete random variable. We use a categorical distribution with five bins to approximate the proposal $\mathsf{q}\left(\mathbf{x}_n|\eta\right)$. The five outputs of the neural network are the probabilities $p_{1:3}$ of drawing one of the three nearest clusters, the probability $p_4$ of the remaining $K-3$ existing clusters (so that each one has probability $p_4/(K-3)$), and the probability $p_5$ of sampling a new cluster. If $K < 3$, the prior proposal is used. If $K = 3$, the $p_4$ is set to zero (all other probabilities are re-normalised).

We set $\mathsf{q}\left(\mathbf{x}_n|\eta\right)$ to the softmax output of the neural network. The aim for the cost function is to maximise the likelihood of the family of discrete distributions given training samples from a discrete distribution. We set the cost function to be the negative log probability given in (3.2). This well relates to neural networks with the negative log of the softmax output. Hence, we can use neural network packages out-of-the-box.

[2]The number of *three* closest clusters is arbitrary chosen by us. It would be interesting, in the possible future research, to vary this number and see how it influences the improvements in convergence.

### 3.0.5 Experiments with the DDPMO model

For our experiments, we chose a soccer video dataset (D'Orazio et al., 2009), for which there already exists a human-authored ground truth. It contains many active objects of different colours that move quickly, both with and without occlusions.

We selected two subsequences of frames to form a training dataset (40 frames) and a test dataset (30 frames). Both datasets consist mostly of intensive play with many players on the field. We use a foreground detector in MATLAB (from package `vision`) to process raw frames and extract positions and colour histograms of foreground pixels[3]. To measure the performance, we use and report commonly used performance metrics: the sequence frame detection accuracy (SFDA) for object detection and the average tracking accuracy (ATA) for tracking (Kasturi et al., 2009).

At first, we run several iterations of sequential Monte Carlo inference in Anglican for this model given the input frames from the training dataset, with 5000 particles. This allows us to extract inputs and outputs for the neural network, as described in the previous section. Then we train the neural network[4] using this extracted data.

Once the neural network is trained, we run inference again both on train and test frame sequences. We measure inference performance with three different types of proposals:

1. the DDPMO prior proposal (i.e. just following the generative model).

2. the data-driven proposal with a trained neural network that outputs probabilities

[3] We trained the foreground detector on frames that did not contribute to the test dataset frame sequence.

[4] We used a feedforward neural network: one hidden layer with 100 nodes, tansig transfer function from the input to the hidden layer, softmax transfer function to the output, and cross-entropy error.

$p_1, \ldots, p_5$ given the observation and the current state of the model during inference.

3. and the data-driven proposal with fixed, hand-tuned probabilities $p_1, \ldots, p_5$. They approximate the distribution over the outputs "1", "2", "3", "any other cluster" and "new" with some smoothing.

We vary number of particles in order to understand how object recognition and tracking performance varies with different proposal types and different number of particles.

Figures 3.3 and 3.4 illustrate the experimental results. For inference with few particles, we get significant improvement in performance using the data-driven proposal. With respect to the particle log-weight, the SMC inference with 10 particles with the data-driven proposal produces results similar to the results from running SMC with thousands of particles under the prior proposal. Thus, using the data-driven proposal, the inference explores the high-probability regions in the posterior space much faster than otherwise.

With respect to performance metrics, for few particles, the performance of SMC with the data-driven proposal is significantly better in comparison to the SMC with the prior proposal with the same number of particles. However, the improvement is less significant, especially in respect to the SFDA metric. In addition, with many particles, SMC with the prior proposal outperforms SMC with the data-driven proposal.

Also, in general, data-driven proposals with the neural network show the same performance as the data-driven proposal with a hand-tuned discriminative model that always returns fixed $p_{1:5}$. However, for the case of SFDA metric performance on the test dataset, the hand-tuned proposal outformed the data-driven proposal with the neural network.

As mentioned earlier, in cases where there were yet no more than 2 clusters, the prior proposal was used. In all experiments using the data-driven proposals, they were used with probability $p^* = 0.8$; thus, with probability $1-p^* = 0.2$, the original prior proposal was used. This mixture proposal probability $p^*$ was incorporated into the particle log-weights, to ensure that that mixture proposal is a valid SMC proposal.

In addition, it is worth noting that even when we attempted to decrease $p^*$ (thus increasing the probability of using the prior proposal), the SFDA metric values for SMC, with the data-driven proposals with 100 particles and more, did not become better for the test dataset and remained very similar to what we see in Figure 3.4. This might mean that, even though the data-driven proposal allows inference to find high-probability posterior regions much faster and with much less computation effort (as shown in "Log-weight" subfigure in Figure 3.4), it is not necessarily the case that all performance metrics of interest will be high for samples from those high-probability posterior regions. On the other hand, the last statement is apparent since the generative model is always only a simplification of the real process. Future experiments might be helpful to provide more experimental details on this.

Scatter plots in Figures 3.5, 3.6, 3.7, 3.8, 3.9, and 3.10 compare the inference time versus particle log-weights, SFDA and ATA metric values.

Examples of frames with detected and tracked objects are provided in Figure 3.11.

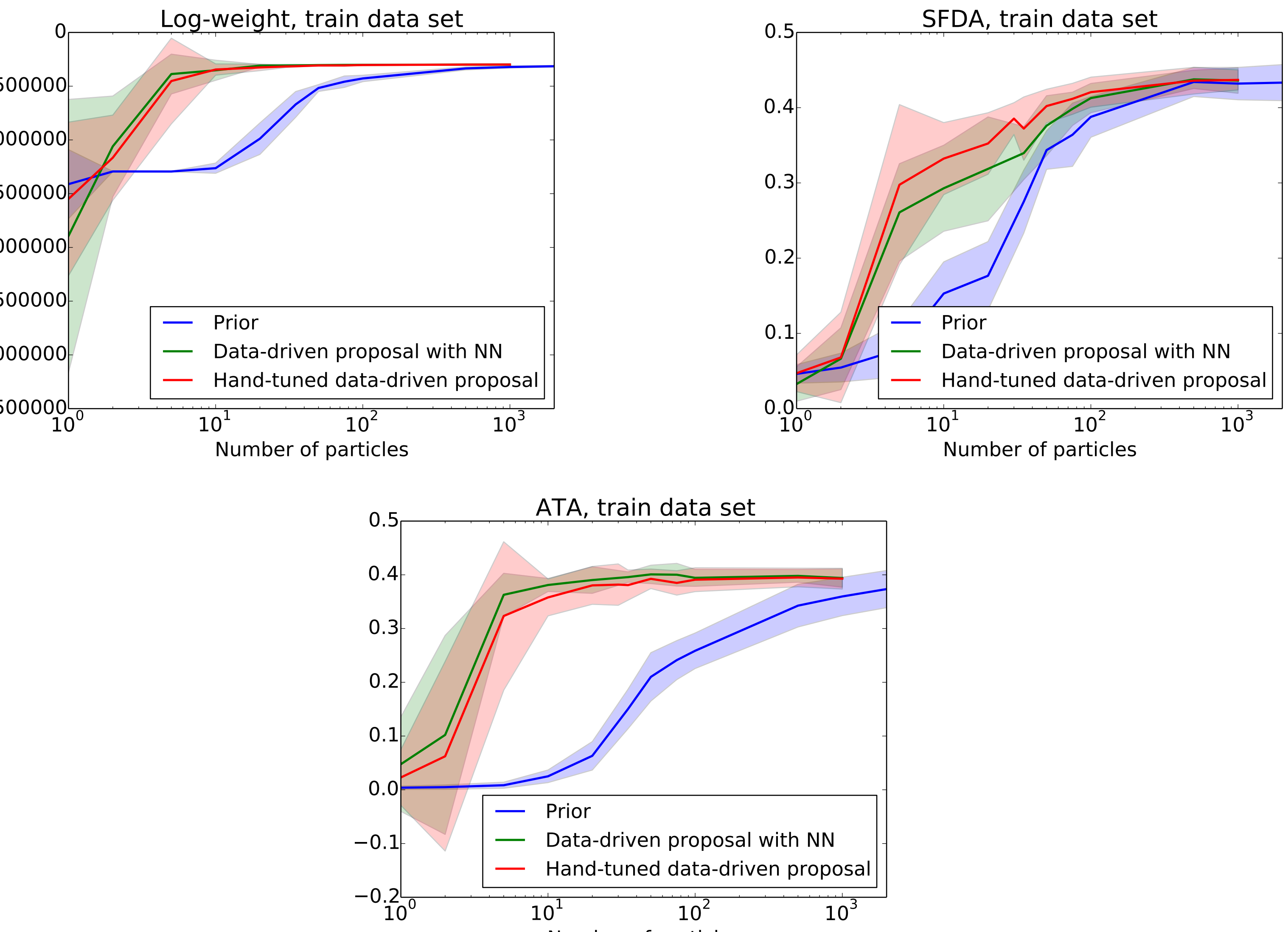

Figure 3.3: Train dataset. Particle log-weight and performance metrics values, namely the sequence frame detection accuracy (SFDA) for object detection and the average tracking accuracy (ATA), for inference results with different proposal types and different number of particles. For the log-weight and both metrics, the higher value is generally better.

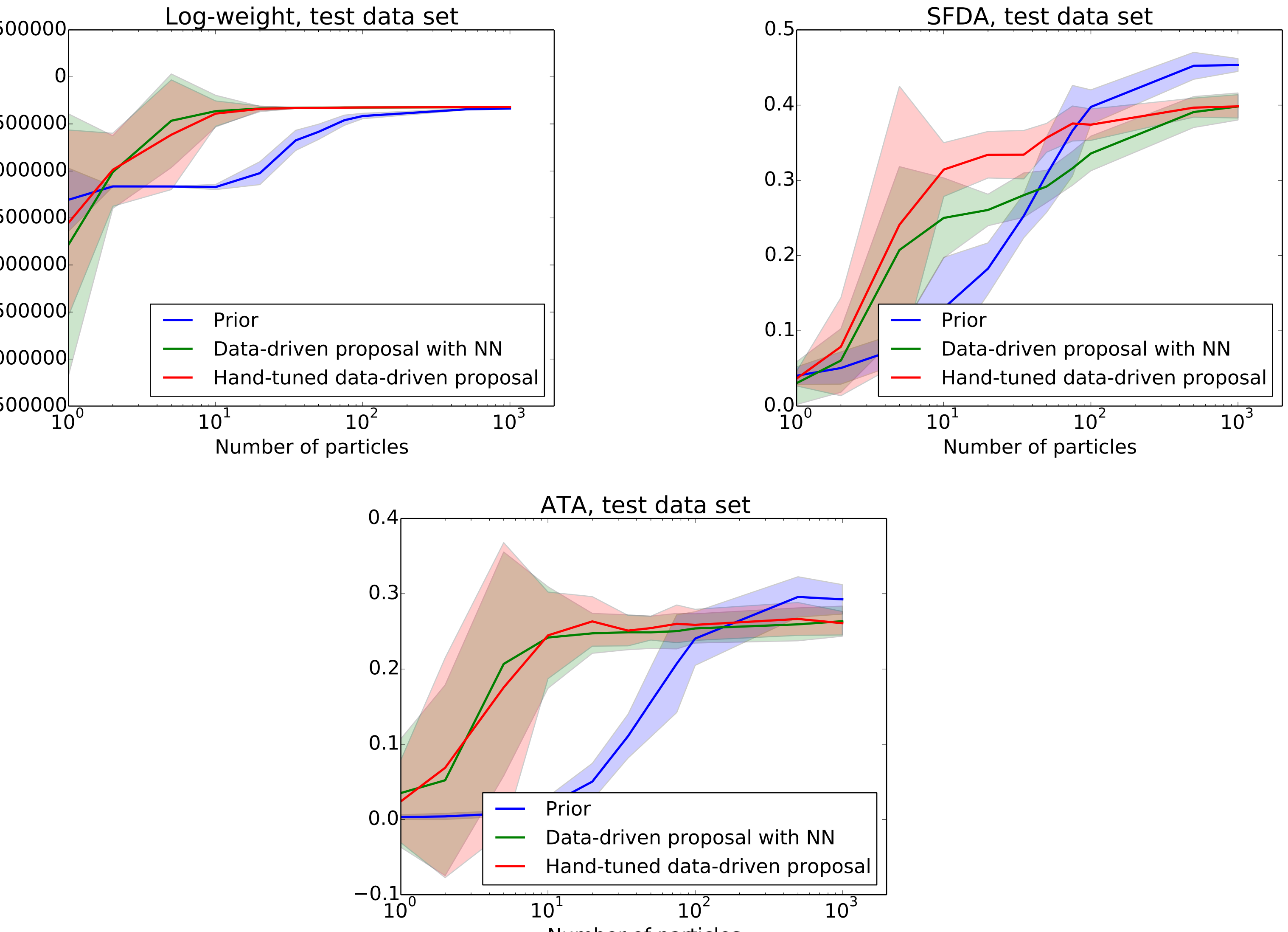

Figure 3.4: Test dataset. Particle log-weight and performance metrics values, namely SFDA and ATA, for inference results with different proposal types and different number of particles. For the log-weight and both metrics, the higher value is generally better.

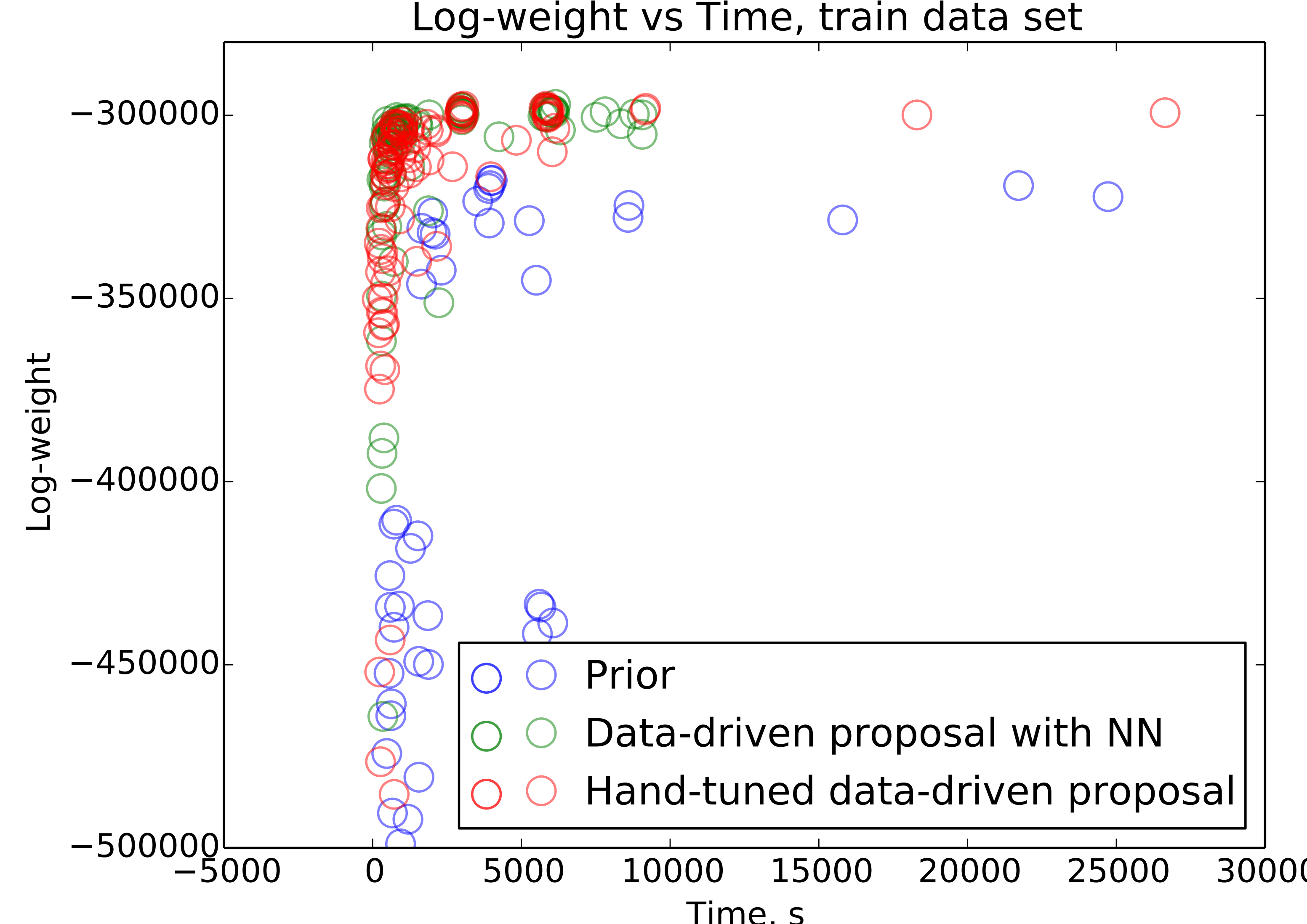

Figure 3.5: Train dataset. The scatter plot that shows how the particle log-weight is affected by the time that inference took (we varied number of particles, thus the time here is a derivative of the number of particles used for the SMC inference). The higher value is generally better.

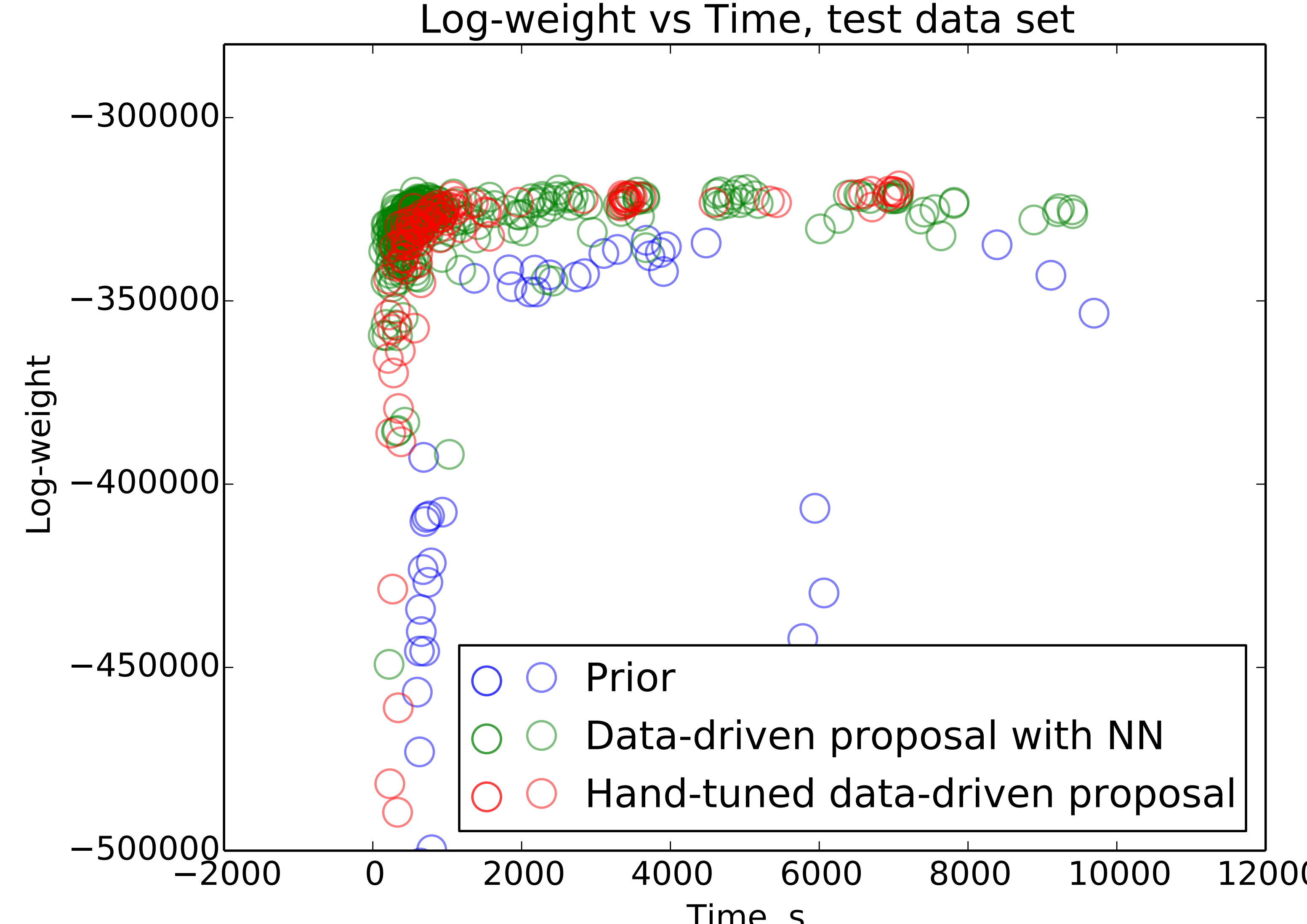

Figure 3.6: Test dataset. The scatter plot that shows how the particle log-weight is affected by the time that inference took (we varied number of particles, thus the time here is a derivative of the number of particles used for the SMC inference). The higher value is generally better.

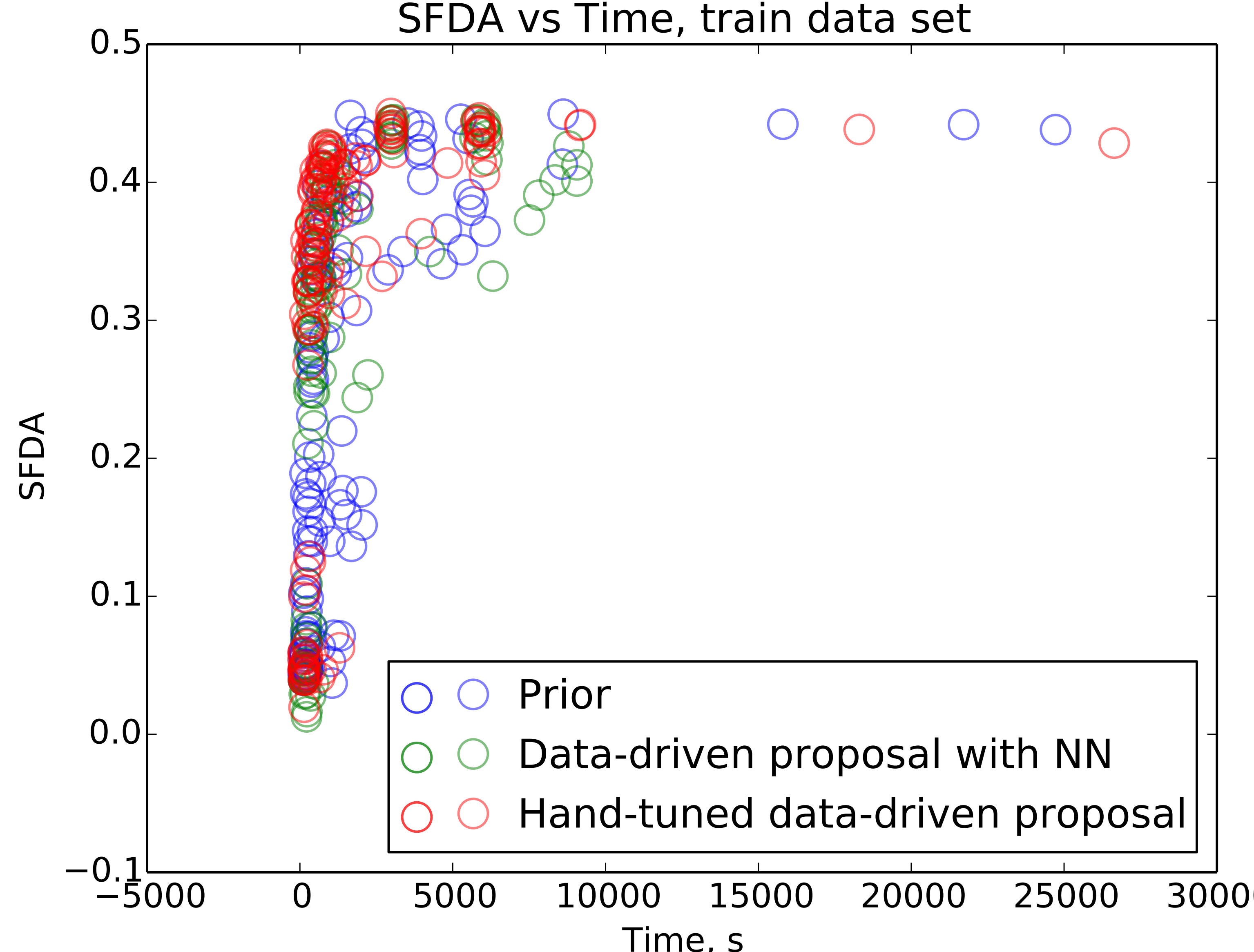

Figure 3.7: Train dataset. The scatter plot that shows how the SFDA metric performance is affected by the time that inference took (we varied number of particles, thus the time here is a derivative of the number of particles used for the SMC inference). The higher value is generally better.

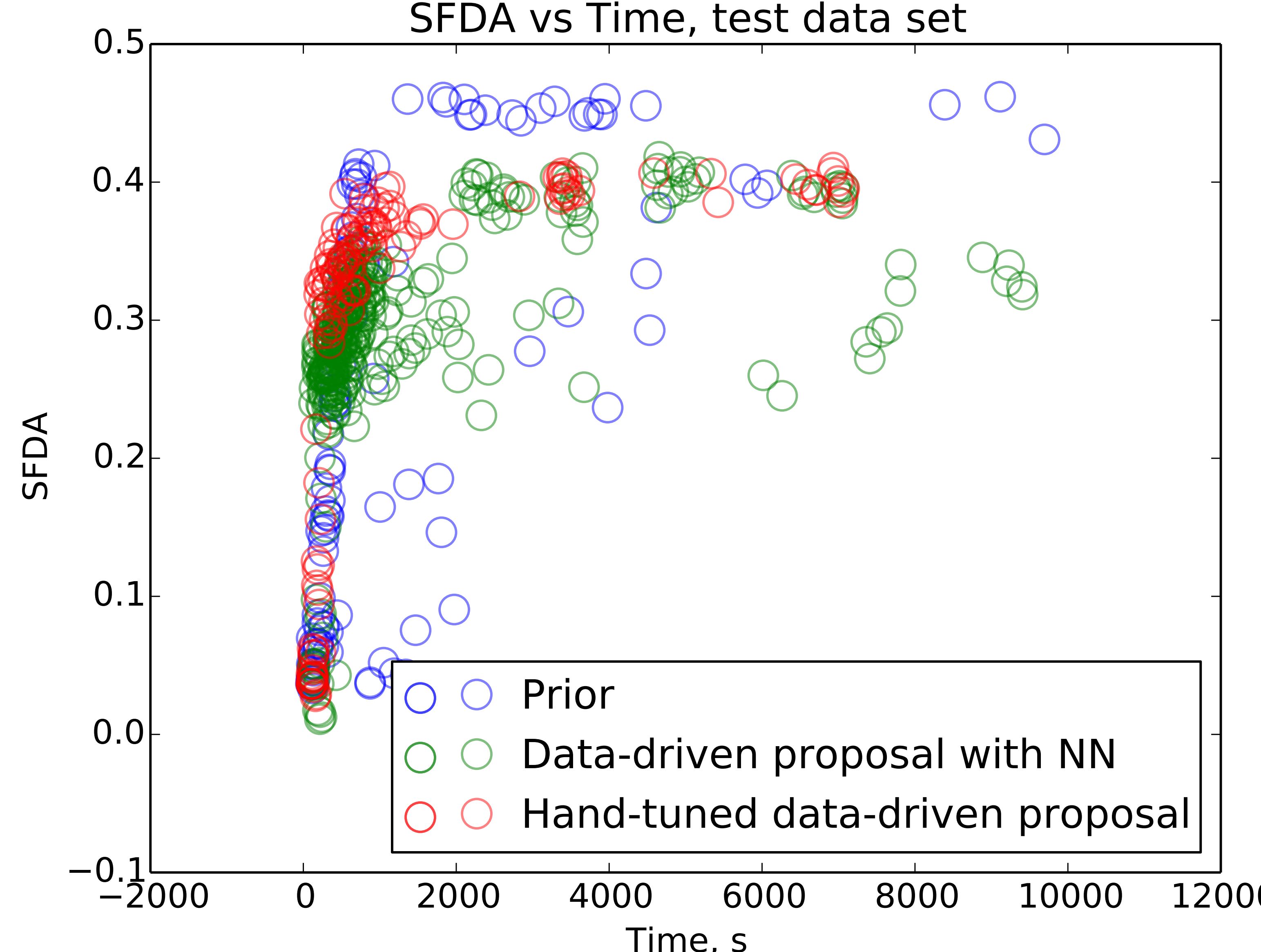

Figure 3.8: Test dataset. The scatter plot that shows how the SFDA metric performance is affected by the time that inference took (we varied number of particles, thus the time here is a derivative of the number of particles used for the SMC inference). The higher value is generally better.

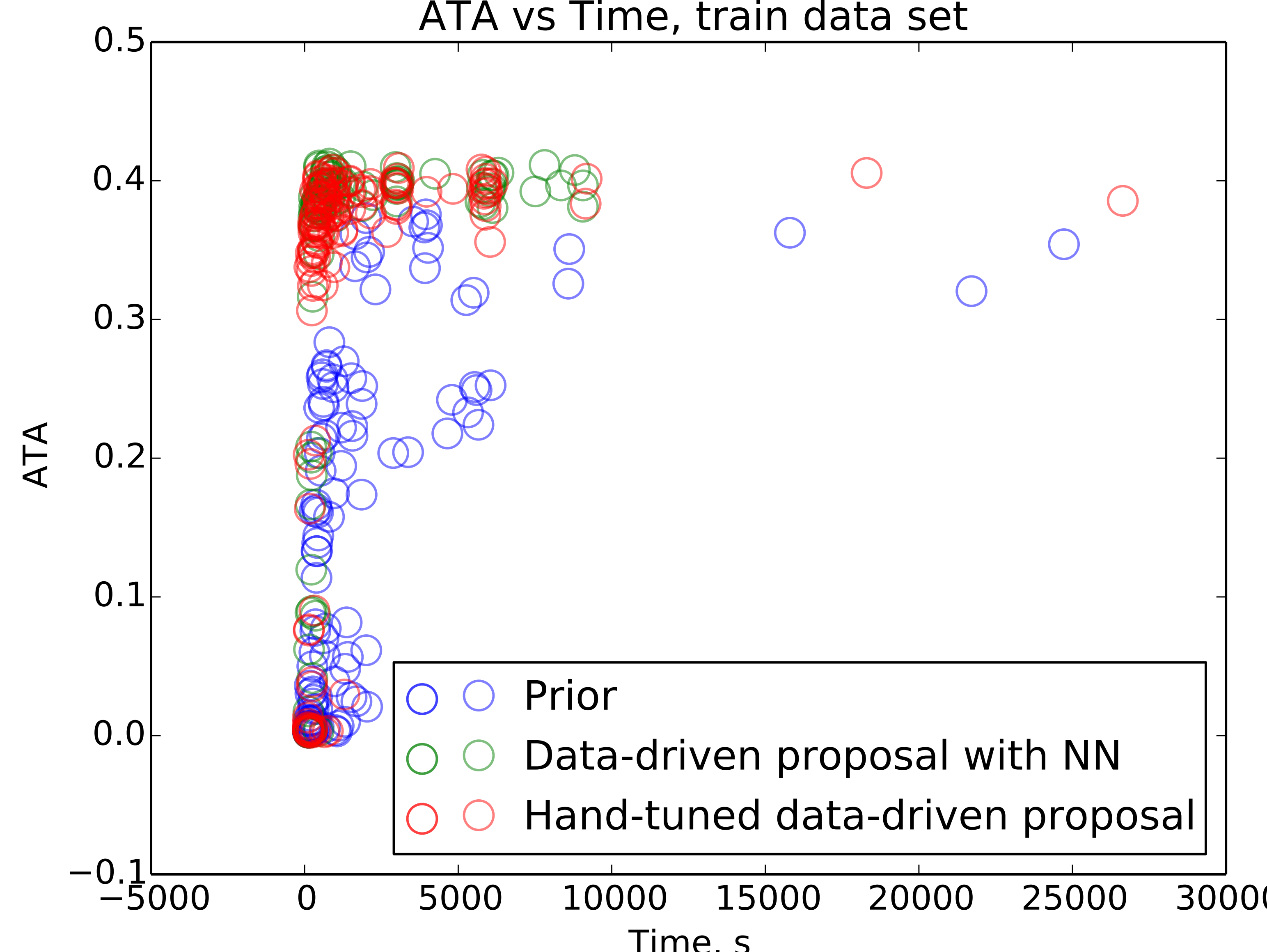

Figure 3.9: Train dataset. The scatter plot that shows how the ATA metric performance is affected by the time that inference took (we varied number of particles, thus the time here is a derivative of the number of particles used for the SMC inference). The higher value is generally better.

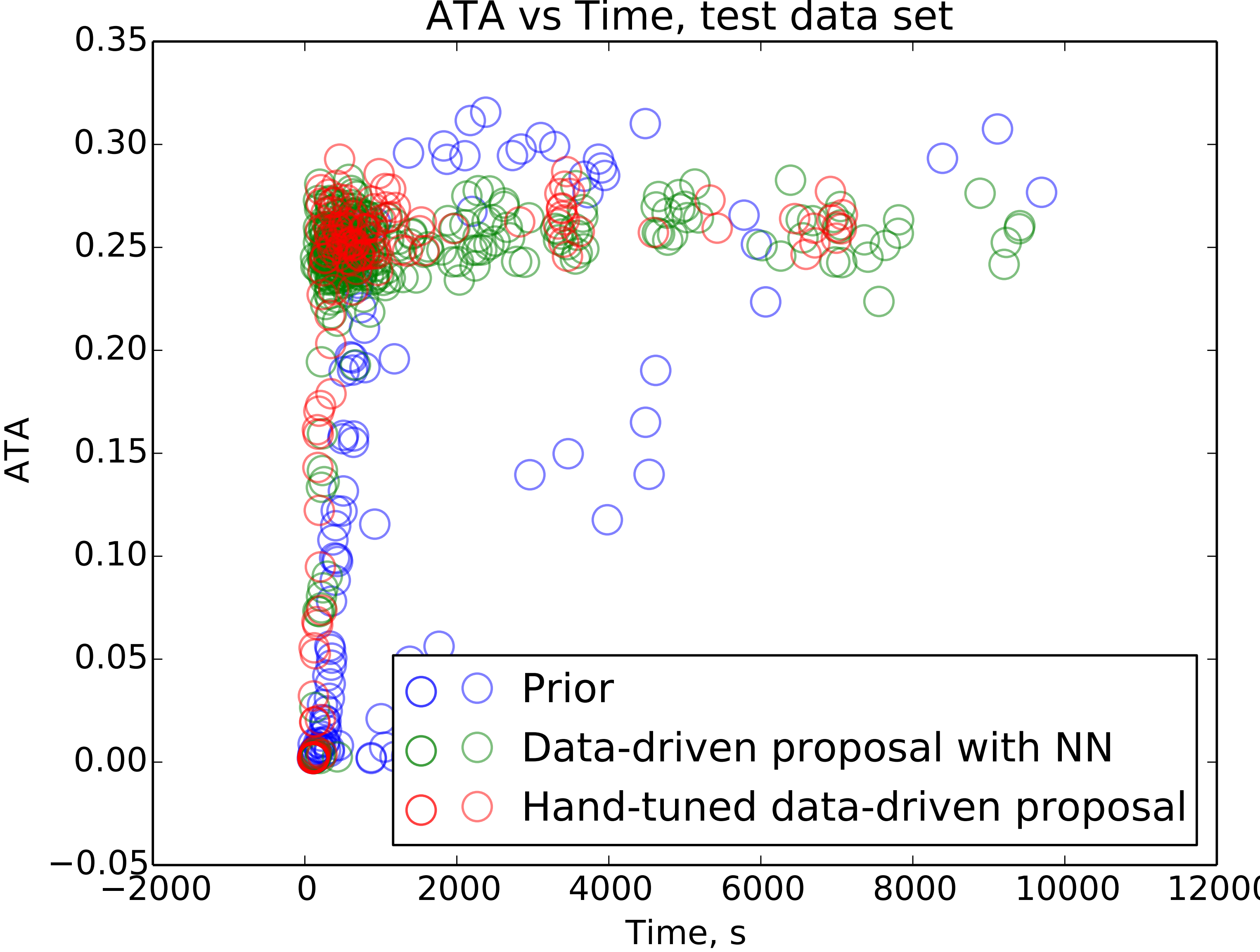

Figure 3.10: Test dataset. The scatter plot that shows how the ATA metric performance is affected by the time that inference took (we varied number of particles, thus the time here is a derivative of the number of particles used for the SMC inference). The higher value is generally better.

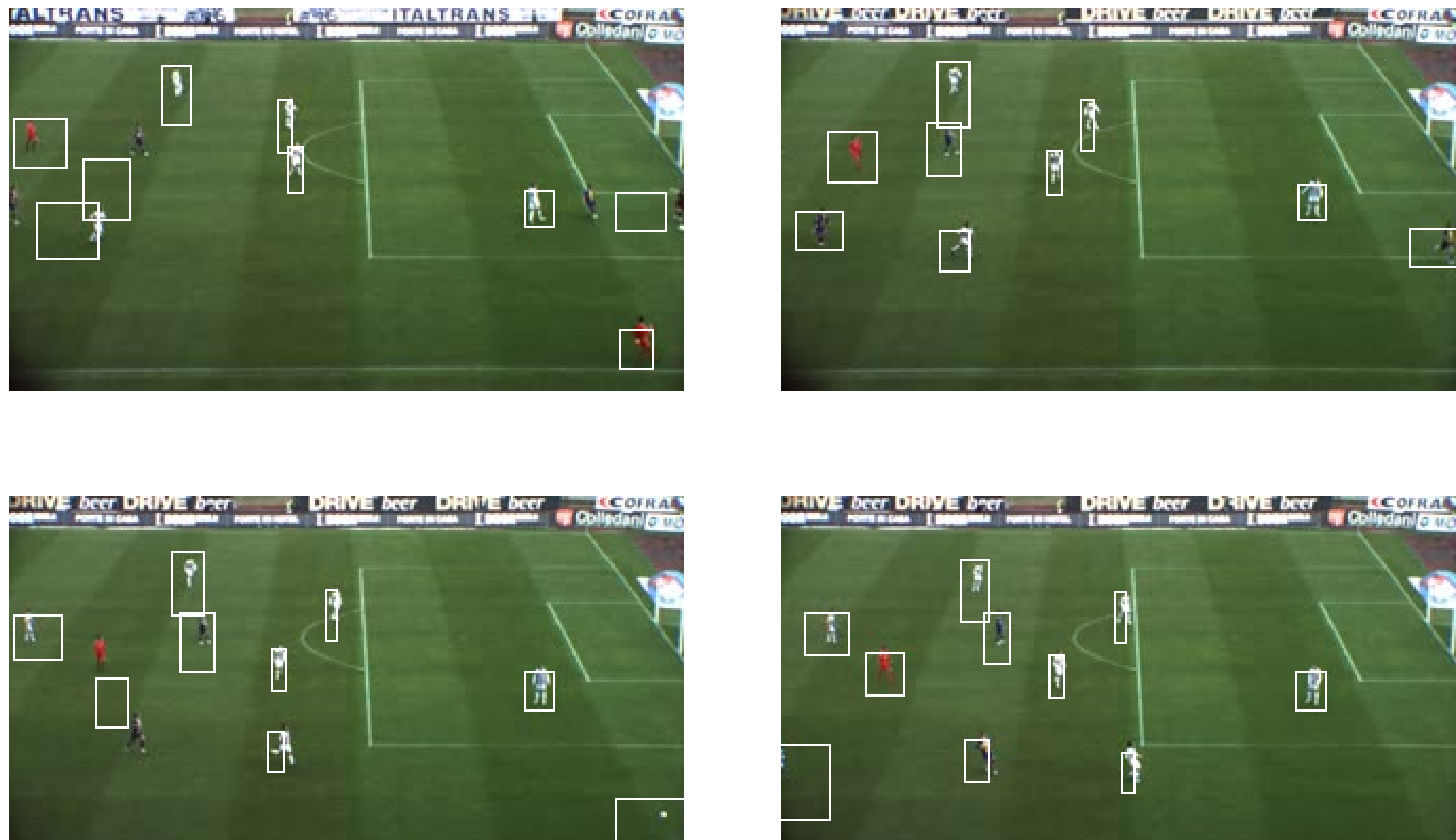

Figure 3.11: An examples of soccer video dataset with detected and tracked objects using DDPMO model in probabilistic programming system Anglican.

## 3.1 Conclusion

Most of inference in currently existing probabilistic programming systems uses prior proposals. Problem-specific proposals are, however, essential for inference, especially in the case of real world applications. New generative models for Bayesian inference are usually introduced, as a conference paper, with a hand-designed proposal that makes the inference in that model feasible.

In this Chapter we presented work on developing a hand-designed data-driven proposal for a particular model, the DDPMO, and directly implementing it in the probabilistic programming system Anglican. Our experimental results showed that the data-driven proposal significantly improves the inference performance so that significantly less particles are necessary to perform good posterior estimation. In general, we assume that our proposal may be applied to any non-parametric generative model with some distance function between clusters and data points (i.e. observations). We performed our experiments in offline settings. To tune the parameters of the proposal, we used neural networks with clearly separated train and test datasets.

The data-driven proposal that we described relies on the feature extractor. The feature extractor maps the current state of the unbounded number of clusters with their sufficient statistics to the input of the neural network. The feature extractor that we implement and use is also a significant part of the data-driven proposal. This is shown by the fact that the neural network performs as well as the fixed hand-tuned discriminative model. This is probably because for the football dataset the spatial factor is very important for the model. Therefore, there is future work to verify whether for more complex datasets and models data-driven proposals with neural networks provide more benefits.

Our work relates to other work in the field on data-driven proposals. The work on using discriminative proposals for Markov Chain Monte Carlo in parametric generative models include (Tu and Zhu, 2002) and (Jampani et al., 2015b), with applications in computer vision. Recent work with sequential Monte Carlo includes neural adaptive SMC (Gu et al., 2015), where authors also adapt proposals by descending the inclusive Kullback-Leibler divergence between the proposal and the true posterior distributions on hidden variables given observations. They use recurrent neural networks to train proposal distributions for inference in parametric generative models with fixed dimensionality. Another related recent work is a new probabilistic programming language called Picture (Kulkarni et al., 2015), for which authors propose and describe the use of data-driven proposals in the context of models for computer vision. They also use neural networks to learn proposals. To get the data to train the neural network, they sample both hidden variables and observations from the generative model unconditionally offline.

In future work, more advanced neural network architectures can be applied to improve results by extracting better features and processing them more efficiently. In particular, one can think of using convolutional neural networks that include as a feature a window of the frame, centered at the new observing pixel. The ultimate goal is to find a way to generate such data-driven proposals automatically, given a generative model in the form of a probabilistic program.

# Conclusion

Probabilistic programming provides users with a favourable way to write probabilistic generative models and perform automatic inference in them. In this work, we explored two applications of it. The first application, learning one-dimensional sampler code, is ultimately directed toward automatic induction of generative probabilistic programs given some examples. The second application, the facilitation of Bayesian inference in probabilistic programming using data-driven proposals, to the development of which we partly contributed, should provide users with much faster inference.

The speed and asymptotics of inference is one of the most important factors for the probabilistic programming field to be successful. There already exist many models written as probabilistic programs, since it is indeed easy to write a directed generative model as a program. However, for many models, even for modest amounts of data, the inference is not tractable with current inference techniques provided in probabilistic programming systems. This means that a significant amount of future work on optimising inference is essential. Ultimately, fast general-purpose inference might also make the automatic learning of complex generative models possible.

# Appendix A

# Corpus of sampler code

```
1 (lambda (p)
2   (if (< (safe-uc 0.0 1.0) p)
3     1.0
4     0.0))
```

Figure A.1: A sampler from the $\mathrm{Bernoulli}(p)$ distribution.

```
1 (lambda (p)
2   (begin
3     (define inner-loop
4       (lambda (voidarg val stack_level)
5         (if (< (safe-uc 0.0 1.0) p)
6           val
7           (if (< stack_level 8)
8             (recur 0.0 (inc val) (inc stack_level)) 0)))))
9     (inner-loop 0.0 1.0 0)))
```

Figure A.2: A sampler from the $\mathrm{Geometric}(p)$ distribution. (Approximately since there is a limiting condition on the stack depth.)

```
(lambda (stack_level)
  (begin
   (define x (safe-uc -1.0 1.0))
   (begin
    (define y (safe-uc -1.0 1.0))
    (begin
     (define s (+ (* x x) (* y y)))
     (if (< s 1.0)
       (* x (safe-sqrt
              (* -2.0 (safe-div (safe-log s) s))))
       (if (< stack_level 8)
         (recur (inc stack_level))
         0)))))))
```

Figure A.3: A sampler from the standard $\text{Normal}(\mu, \sigma)$ distribution. (Approximately since there is a limiting condition on the stack depth.)

```
(lambda (mean std)
  (begin
   (define x (safe-uc -1.0 1.0))
   (begin
    (define y (safe-uc -1.0 1.0))
    (begin
     (define s (+ (* x x) (* y y)))
     (if (< s 1.0)
       (+ mean
          (* std (* x (safe-sqrt
                        (* -2.0 (safe-div
                                  (safe-log s) s)))))))
       (if (< stack_level 8)
         (recur (inc stack_level))
         0)))))))
```

Figure A.4: A sampler from the general $\text{Normal}(0, 1)$ distribution. (Approximately since there is a limiting condition on the stack depth.)

```
1  (lambda (rate)
2    (begin
3     (define L (exp (* -1.0 rate)))
4     (begin
5      (define inner-loop
6        (lambda (k p stack_level)
7                (if (< p L)
8                  (dec k)
9                  (begin
10                  (define u (safe-uc 0.0 1.0))
11                  (if (< stack_level 8)
12                    (recur (inc k) (* p u)
13                          (inc stack_level)) 0)
14                  ))))
15     (inner-loop 1.0 (safe-uc 0.0 1.0)))))
```

Figure A.5: A sampler from the $\mathrm{Poisson}(\lambda)$ distribution. (Approximately since there is a limiting condition on the stack depth.)

```
(lambda
  (alpha stack_level)
  (if (< (safe-uc 0.0 1.0)
         (safe-div (exp 1.0)
                   (+ (exp 1.0) alpha)))
    (begin
      (define epsilon
        (exp (* (safe-div 1.0 alpha)
                (safe-log (safe-uc 0.0 1.0)))))
      (if (< (exp (* -1.0 epsilon))
             (safe-uc 0.0 1.0))
        (if (< stack_level 8)
          (recur (inc stack_level))
          0)
        epsilon))
    (begin
      (define epsilon
        (- 1.0
           (safe-log (safe-uc 0.0 1.0))))
      (if (< (exp (* (dec alpha)
                     (safe-log epsilon)))
             (safe-uc 0.0 1.0))
        (if (< stack_level 8)
          (recur (inc stack_level))
          0)
        epsilon))))
```

Figure A.6: A sampler from the $\mathrm{Gamma}(\alpha, 1.0)$ distribution. (Approximately since there is a limiting condition on the stack depth.)

```
(lambda (alpha beta)
  (begin
   (define X
     (begin
      (define get-gamma-1
        (lambda (void1 void2 stack_level)
                (if (< (safe-uc 0.0 1.0)
                       (safe-div (exp 1.0) (+ (exp 1.0) alpha)))
                  (begin
                   (define epsilon (exp (* (safe-div 1.0 alpha)
                                           (safe-log (safe-uc 0.0 1.0)))))
                   (if (< (exp (* -1.0 epsilon)) (safe-uc 0.0 1.0))
                     (if (< stack_level 8) (recur 0.0 0.0 (inc stack_level)) 0)
                     epsilon))
                  (begin
                   (define epsilon (- 1.0 (safe-log (safe-uc 0.0 1.0))))
                   (if (< (exp (* (dec alpha) (safe-log epsilon))) (safe-uc 0.0 1.0))
                     (if (< stack_level 8) (recur 0.0 0.0 (inc stack_level)) 0)
                     epsilon)))))
      (get-gamma-1 0.0 0.0 0)))
   (begin
    (define Y
      (begin
       (define get-gamma-2
         (lambda (void1 void2 stack_level)
                 (if (< (safe-uc 0.0 1.0) (safe-div (exp 1.0) (+ (exp 1.0) alpha)))
                   (begin
                    (define epsilon (exp (* (safe-div 1.0 alpha)
                                            (safe-log (safe-uc 0.0 1.0)))))
                    (if (< (exp (* -1.0 epsilon)) (safe-uc 0.0 1.0))
                      (if (< stack_level 8) (recur 0.0 0.0 (inc stack_level)) 0)
                      epsilon))
                   (begin
                    (define epsilon (- 1.0 (safe-log (safe-uc 0.0 1.0))))
                    (if (< (exp (* (dec alpha) (safe-log epsilon))) (safe-uc 0.0 1.0))
                      (if (< stack_level 8) (recur 0.0 0.0 (inc stack_level)) 0)
                      epsilon)))))
       (get-gamma-2 0.0 0.0 0)))
    (safe-div X (+ X Y)))))
```

Figure A.7: A sampler from the $\mathrm{Beta}(\alpha, \beta)$ distribution. (Approximately since there is a limiting condition on the stack depth.)

```
(lambda (alpha)
  (exp (safe-div
        (safe-log (safe-uc 0.0 1.0)) alpha)))
```

Figure A.8: A sampler from the $\mathrm{Beta}(\alpha, 1)$ distribution.

# Appendix B

# The DDPMO and GPU code in Anglican

## B.1 The DDPMO code

```
(ns ddpmo.ddpmo
  (:use [anglican emit runtime]
        [anglib xrp utils new-dists anglican-utils]
        ddpmo.ddpmo-header)
  (:require [clojure.core.matrix :as m]
            [clojure.core.matrix
             :refer [identity-matrix mmul add sub transpose
                     matrix to-nested-vectors]
             :rename {identity-matrix eye
                      add madd
                      sub msub
                      transpose mtranspose}]
            [clojure.core.matrix.linear :as ml]))

(with-primitive-procedures
  [multivariate-t mvn-conjugate-fast
   dirichlet-multinomial-process
   DIRICHLET-MULTINOMIAL-PROCESS-STATE-INFO
   MVN-PROCESS-FAST-STATE-INFO
   matrix produce-matrix-from-vector to-nested-vectors
   mtranspose matrix-to-clojure-vector]

  (defquery ddpmo
    "The Dependent Dirichlet Process Mixture of Objects
    for Detection-free Tracking"
```

```
26      [data Nts proposal-type]
27
28      (let [
29
30              ;;;;;; DDPMO model ;;;;;;
31
32              ;; Hyperparameters for squares/objects/football
33              alpha 0.1 ; for GPU
34              rho 0.32 ; for GPU
35              mu-0 (produce-matrix-from-vector [0 0]) ; for NiW
36              k-0 0.00370790649926 ; for normal-inverse-wishart
37              nu-0 7336.3104796 ; for normal-inverse-wishart
38              Lambda-0 (matrix [[193.362493995 0]
39                                [0 40.6543682123]])
40              q-0 (vec (repeat 10 10.0)) ; for Dirichlet.
41              ;; The dimensionality must
42              ;; match number of RGB bins V
43              M 10.1 ; for G0 (eqns (7-8))
44              multinomial-trials 49 ; for eqn (2)...
45              ;; this is m x m where m = 2L + 1
46
47              extract-old-style-theta
48              (fn [theta]
49                (let
50                  [mvn-process (retrieve (get theta 'positions))
51                   dirichlet-multinomial-process-instance
52                   (retrieve (get theta 'colours))
53                   mu-Sigma (MVN-PROCESS-FAST-STATE-INFO mvn-process)
54                   ps (DIRICHLET-MULTINOMIAL-PROCESS-STATE-INFO
55                        dirichlet-multinomial-process-instance)
56                   theta]
57                  {'mu (get mu-Sigma 'mu) 'Sigma
58                   (get mu-Sigma 'Sigma)
59                   'trials multinomial-trials 'ps ps}))
60
61              get-N (fn [t] (nth Nts (dec t)))
62
63              ;; Transition distribution
64              T (fn T [prev-theta]
65                  (let [previous-mvn-process
66                        (get prev-theta 'positions)
67                        previous-dirichlet-multinomial-process
68                        (get prev-theta 'colours)
```

```
 69
 70                     new-mvn-process
 71                     (XRP (mvn-conjugate-fast
 72                            mu-0 k-0 nu-0 Lambda-0))
 73                     new-dirichlet-multinomial-process
 74                     (XRP
 75                       (dirichlet-multinomial-process
 76                         q-0 multinomial-trials))
 77
 78                     ;; Auxiliary transition
 79                     _ (repeatedly
 80                         M (fn []
 81                             (INCORPORATE
 82                               new-mvn-process
 83                               (SAMPLE
 84                                 previous-mvn-process))))
 85                     _ (repeatedly
 86                         M
 87                         (fn []
 88                           (INCORPORATE
 89                             new-dirichlet-multinomial-process
 90                             (SAMPLE
 91                               previous-dirichlet-multinomial-process))))
 92                     ]
 93                 {'positions new-mvn-process
 94                  'colours new-dirichlet-multinomial-process}))
 95
 96           ;; Base distribution
 97           G0 (fn G0 []
 98                (let [mvn-process
 99                      (XRP
100                        (mvn-conjugate-fast
101                          mu-0 k-0 nu-0 Lambda-0))
102                      dirichlet-multinomial-process-instance
103                      (XRP
104                        (dirichlet-multinomial-process
105                          q-0 multinomial-trials))
106                      ]
107                  {'positions mvn-process
108                   'colours dirichlet-multinomial-process-instance}))
109
110           [gpu get-theta] (create-gpu alpha rho G0 T get-N)
111
```

```
112  ;; Helper function
113  ;; Returns parameters for the corresponding
114  ;; table of foreground pixel n at time t
115  get-theta-t-n (mem (fn get-theta-t-n [t n]
116                       (let [customers (gpu t n)
117                             cs (get customers 'cs)
118                             k (get cs (dec n))]
119                         (get-theta t k))))
120
121  ;;;;;; OBSERVES ;;;;;;
122  observe-lines
123  (fn observe-lines [lines line-id]
124    (if (nil? (first lines))
125      true
126      (let [line (first lines)
127            pos (get line 'pos)
128            _ (store "current-pos"
129                     (matrix-to-clojure-vector pos))
130            col (get line 'col)
131            _ (store "current-col" col)
132            t (get line 't)
133            n (get line 'n)
134            theta (get-theta-t-n t n)
135            positions-process (get theta 'positions)
136            colours-process (get theta 'colours)]
137
138        ; Observing positions.
139        (OBSERVE positions-process pos)
140
141        ; Observing colours
142        (OBSERVE colours-process col)
143
144        (if (= n (get-N t))
145          (let [gpu (gpu t n)
146                cs (get gpu 'cs)
147                relevant-clusters (distinct cs)
148                thetas
149                (map (fn [k]
150                       (let [theta (get-theta t k)
151                             theta
152                             (extract-old-style-theta theta)
153                             mu (get theta 'mu)
154                             Sigma (get theta 'Sigma)
```

```
155                                       ps (get theta 'ps)]
156                                   {'k k 'mu mu 'Sigma Sigma 'ps ps}))
157                               relevant-clusters)
158                          res {'t t 'n n 'gpu gpu 'thetas thetas}]
159                      (predict res)))
160                  (observe-lines (rest lines) (inc line-id))))))]
161
162       (observe-lines data 0))))
163
164 (defn -main [data-set-name number-of-particles
165              num-particles-to-output
166              proposal-type & ignore-following-args]
167   (let [number-of-particles (parse-int number-of-particles)
168         num-particles-to-output
169         (parse-int num-particles-to-output)
170         proposal-type (str proposal-type)
171         _ (case proposal-type "prior"
172             :okay "handtuned" :okay "nn" :okay)
173         [data Nts] (load-DDPMO-data data-set-name)
174         query-results (doquery
175                         :smc ddpmo [data Nts proposal-type]
176                         :number-of-particles number-of-particles)
177         results
178         (doall
179           (map
180             (fn [particle-output particle-id]
181               (doall
182                 (map
183                   (fn [x]
184                     (println
185                       (str particle-id ","
186                            (first x) "," (second x) ",0.0")))
187                   (get particle-output :anglican.state/predicts))))
188             (take num-particles-to-output query-results)
189             (range num-particles-to-output)))]
190     results))
```

## B.2 The GPU code

```
1 ;;;;;; GPU definition ;;;;;;
2
3 ; Creates an instance of a GPU process.
```

```
4   ; Takes:
5   ; * GPU's alpha and rho.
6   ; * Base distribution G0.
7   ; * Transition distribution T.
8   ; * function get-N which returns the number
9   ; of points at each time.
10  ; Returns: [gpu get-theta]
11  (defm create-gpu [alpha rho G0 T get-N]
12    (let
13      [;; Given vector of table sizes ms = [m1 m2 ...],
14        ;; returns a new vector of table
15        ;; sizes by removing customers from tables
16        ;; with probability rho
17        remove-customers
18        (fn [ms]
19          (vec (map (fn [m]
20                      (if (= m 0)
21                       0
22                        (- m
23                           (SAMPLE
24                             (binomial m rho)))))
25                  ms)))
26
27        ;; Returns {'cs (vector of n cluster ids)
28        ;;          'K (number of unique clusters at n)
29        ;;          'ms (vector of cluster sizes at n)}
30        ;; after processing foreground pixel n at time t
31        ;; n goes from 1
32        ;; c_i goes from 0
33        ;; K = max(c_i) + 1
34        ;; t goes from 1
35        gpu (mem
36              (fn gpu [t n]
37                (if (= n 0)
38                  ;; Initialise
39                  (if (= t 1)
40                    {'cs '[] 'K 0 'ms '[]}
41                    (let [prev-t-gpu (gpu (dec t) (get-N (dec t)))
42                          prev-K (get prev-t-gpu 'K)
43                          prev-ms (get prev-t-gpu 'ms)]
44                      {'cs '[] 'K prev-K
45                       'ms (remove-customers prev-ms)}))
46
```

```
                ;; Get from step (n - 1)
                (let [prev-n-gpu (gpu t (dec n))
                      cs (get prev-n-gpu 'cs)
                      K (get prev-n-gpu 'K)
                      ms (get prev-n-gpu 'ms)
                      w (conj ms alpha)
                      c (SAMPLE (discrete w))
                      new-cs (conj cs c)
                      new-K (max K (inc c))
                      new-ms (assoc ms c (inc (get ms c 0)))]
                  {'cs new-cs 'K new-K 'ms new-ms}))))

      ;; Returns parameters for table k at time t
      ;; using either transition distribution T
      ;; or base distribution G0
      get-theta (mem (fn get-theta [t k]
                       (if (= t 1)
                         (G0)
                         (let [prev-customers
                               (gpu (dec t) (get-N (dec t)))
                               prev-K (get prev-customers 'K)
                               initial-ms (get (gpu t 0) 'ms)]
                           (if (> k (dec prev-K))
                             (G0)
                             (if (= (nth initial-ms k) 0)
                               nil
                               (T (get-theta (dec t) k)))))))))]
    [gpu get-theta]))
```

## B.3 Clojure code for the data-driven proposal

```
(def NUMBER-OF-NEAREST-CLUSTERS 3)

(def sort-thetas
  (fn [thetas]
    (let
      [my-comparer
       (fn [el1 el2]
         (< (nth el1 2) (nth el2 2)))]
      (sort my-comparer thetas))))

(def distance
```

```
  (fn [[x1 y1] [x2 y2]]
    "Returns Euclidean distance between two 2D points."
    ;; Important! Here x is really y, and vice versa.
    ;; This is because in the MATLAB code the first
    ;; coordinate is y.
    (pow (+ (pow (- x1 x2) 2.0) (pow (- y1 y2) 2.0)) 0.5)))
```

## B.4 Anglican code (within the DDPMO model) for the data-driven proposal

```
get-thetas
(fn [t n]
  "Returns thetas for active clusters (ms[i] > 0)
  at data point (t, n). This function should be
  called only when we already processed that data point."
  (let
    [
     gpu-state (gpu t n)
     ms (get gpu-state 'ms)
     get-theta (fn [t k] (if (> (nth ms k) 0)
                           (get-theta t k)
                           nil))
     thetas (map (fn [k] (list k (get-theta t k)))
                 (range (count ms)))
     thetas (filter
              (fn [el] (not (nil? (second el)))) thetas)
     ]
    thetas))

get-mean-coords
(fn [theta]
  "Extracts mean from the theta as Clojure vector."
  (let
    [coords (matrix-to-clojure-vector
               (get (MVN-PROCESS-FAST-STATE-INFO
                       (retrieve (get theta 'positions)))
                    'mu))]
    coords))

get-nearest-thetas
(fn [t n [x y]]
  "Gets an ordered list of theta which are the nearest
```

```
  to the point [x y] based on the state at the previous
  data point (t, n - 1)."
  (if (and (= t 1) (= n 1))
    nil
    (let
      [[t n]
       (if (= n 1)
         [(- t 1) (get-N (- t 1))]
         [t (- n 1)])]
      (let
        [thetas (get-thetas t n)
         thetas (map (fn [[k theta]]
                       (list k theta
                             (distance [x y]
                               (get-mean-coords theta))))
                     thetas)
         thetas (sort-thetas thetas)
         thetas (take NUMBER-OF-NEAREST-CLUSTERS thetas)]
        (if (< (count thetas) NUMBER-OF-NEAREST-CLUSTERS)
          nil
          thetas)))))

;; Do the trick to allow mutual recursion.
_ (store "get-nearest-thetas" get-nearest-thetas)
```

## B.5 Code for the GPU, to get data for the proposal for train datasets

```
NEAREST-THETAS ((retrieve "get-nearest-thetas")
                  t n (retrieve "current-pos"))
for-proposal
  (map
   (fn [the-list]
     (let
       [theta-id (nth the-list 0)
        theta (nth the-list 1)
        distance-to-the-center (nth the-list 2)]
       (list
        theta-id
        (DIRICHLET-MULTINOMIAL-PROCESS-STATE-INFO
          (retrieve (get theta 'colours)))
        distance-to-the-center)))
```

```
    NEAREST-THETAS)
nn-input
   (concat
    (apply concat
      (doall
        (map
          (fn [data]
            (concat (nth data 1)
                    (list (nth data 2))))
          for-proposal)))
    (doall (map (fn [x] (/ x 49.0))
                (retrieve "current-col")))))
c (if (or (not (= (count nn-input) 43))
          (= proposal-type "prior"))
     (SAMPLE (discrete w))
     (let
       [dist (sample-cluster-id
               nn-input w 0.8
               (map first NEAREST-THETAS)
               (= proposal-type "handtuned"))
        [my-sample log-likelihood] (sample dist)]
       (add-log-weight log-likelihood)
       my-sample))
```

## B.6 Code for the GPU, to use the proposal for test datasets

```
NEAREST-THETAS ((retrieve "get-nearest-thetas")
                        t n (retrieve "current-pos"))
for-proposal
   (map
    (fn [the-list]
      (let
        [theta-id (nth the-list 0)
         theta (nth the-list 1)
         distance-to-the-center (nth the-list 2)]
        (list
         theta-id
         (DIRICHLET-MULTINOMIAL-PROCESS-STATE-INFO
           (retrieve (get theta 'colours)))
         distance-to-the-center)))
    NEAREST-THETAS)
_ (predict (list for-proposal c
```

```
17                          (retrieve "current-col")
18                          (count w)))
```